\documentclass{article}

\usepackage{PRIMEarxiv}

\usepackage[utf8]{inputenc} 
\usepackage[T1]{fontenc}    
\usepackage{hyperref}       
\usepackage{url}            
\usepackage{booktabs}       
\usepackage{amsfonts}       
\usepackage{nicefrac}       
\usepackage{microtype}      
\usepackage{lipsum}
\usepackage{setspace}
\usepackage{titlesec}
\usepackage{tocloft}
\usepackage{parskip}
\usepackage{placeins}
\usepackage{gensymb}
\usepackage{amssymb}
\usepackage{amsmath}
\usepackage{float}
\usepackage{biblatex}
\usepackage{newtxtext, newtxmath}
\usepackage{fancyhdr}       
\usepackage{graphicx}       
\graphicspath{{media/}}     

\title{CONTROL OF A PENDULUM SYSTEM: FROM
SIMULATION TO REALITY
}

\author{
  Iyer Venkataraman Natarajan \\
  School of Mechanical and Aerospace Engineering \\
  Nanyang Technological University \\
  Singapore\\
  \texttt{venk0023@e.ntu.edu.sg} 
}

\begin{document}
\maketitle

\begin{abstract}
Control theory deals with the study of controlling dynamical systems. Robots today
are growing increasingly complex and moving out of factory floors to real world
environment. These robots have to interact with real world environment factors such as
disturbances and this requires the robot to have a control system that is robust. Testing
control algorithms on robots in real world environment can pose critical safety issues and
can be financially expensive. This has resulted in a heavy emphasis on using simulation
to test control algorithms before deploying them in real world environments. Designing
control algorithms is an iterative process that starts with modelling the target system in
simulation, designing a controller, testing the controller in simulation and then changing
the controller parameters to design a better controller. This report explores how an
approximated system model of a target hardware system can be developed, which can
then be used to design a LQR controller for the target system. The controller is then
tested under a disturbance, on hardware and in simulation, and the system response is
recorded. The system response from hardware and simulation are then compared to
validate the use of approximated system models in simulation for designing and testing
control algorithms.
\end{abstract}

\section{Introduction}
A robot can be controlled by modelling it as a dynamical system. Modelling such dynamical systems can get increasingly complex when they need to interact with external environment where safety can be of huge importance. Simulation can be a vital tool to model these dynamical systems and test control algorithms before deploying them in real world. 

Control algorithms consist of  \cite{prasad_tyagi_gupta_2014}:
\begin{itemize}
    \item Classical control algorithms
    \item Optimal control algorithms
\end{itemize}

Classical control algorithms have been around since the 1940's. These include control techniques which make use of 
\begin{itemize}
    \item Root locus plots
    \item Bode plots
    \item Routh-Hurwitz criteria for stability
\end{itemize}
These methods employ the use of Laplace transform to convert the system dynamics to the frequency domain and use feedback to change the system dynamics as desired. 

Optimal control algorithms can be employed to MIMO (Multiple Input Multiple Output) systems by making use of the state space representation of an ${n^{th}}$ order dynamical system in the time domain. This has led to powerful control techniques which include 
\begin{itemize}
    \item LQR (Linear Quadratic Regulator)
    \item  MPC (Model Predictive control)
\end{itemize}
that come up with optimised solutions to control problems \cite{state}.

Optimal control algorithms optimise a cost function. The cost function can be a function of the state of the dynamical system, the amount of control input used etc. 
These set of algorithms have proved to be robust and have been shown to perform better than traditional control algorithm like PID \cite{saraf_gupta_alivelu_parimi}.

The advantages of using LQR controller when compared to a standard PID controller are stated below
\begin{itemize}
    \item LQR controllers are more robust and perform better on metrics such as maximum overshoot from desired state. \cite{saraf_gupta_alivelu_parimi}
    \item The gains for the LQR controller do not need tuning, as LQR outputs an optimal gain matrix.
    \item The input to the LQR controller are the cost matrices ${Q}$ and ${R}$, which represent the cost associated with being in a state and using effort to regulate the system at the desired set point respectively. This makes the process of designing a suitable controller easier and more intuitive.
\end{itemize}

Hence, LQR has been used as the control algorithm of choice in this report. In order to design a LQR controller and also model a given dynamical system in simulation, the dynamic parameters of the dynamical system in consideration have to be known. 

A pendulum system is the simplest possible non linear dynamical system. It can be modelled easily and can be used to test a variety of control algorithms. Hence, a pendulum system is used as the dynamical system of choice in this report. A pendulum system is setup using relevant hardware components and a set of experiments are designed to obtain the corresponding system dynamics parameters in order to obtain the approximated system model.

A simulation of the pendulum system is developed using the approximated system model. A suitable LQR controller is designed and is tested in simulation in the presence of a disturbance, modelled in the form of white noise. The system response in simulation is recorded. The same LQR controller is tested on the physical pendulum setup, in the presence of a disturbance, modelled in the form of white noise. The system response is recorded. 

The system response obtained from testing the LQR controller in simulation and on the physical setup is compared and statistical data analysis techniques are used to validate if an approximated system model in simulation can be used to test control algorithms.

The report flow is as below
\begin{itemize}
    \item A look into similar work done in literature
    \item Designing experiments to obtain the approximated system dynamics parameters of the hardware
    \item Setup of simulation to model the hardware dynamics 
    \item Designing a LQR controller and testing it under a disturbance in simulation and on hardware
    \item Analysis of performance of simulation in modelling the hardware response under the controller
\end{itemize}

\textbf{\textit{NOTE:}} \textit{All the plots in the report are in SI units. The usage of the terms Position, Velocity, Acceleration and Effort in plots and text, correspond to Angular position, Angular velocity, Angular acceleration and Torque respectively of the pendulum system. Hence, the units of Position, Velocity, Acceleration and Effort are rad, ${\frac{rad}{s}}$, ${\frac{rad}{s^2}}$ and Nm respectively.}

\section{Literature Review}
A lot of work has been done in the field of designing control algorithms for dynamical systems and testing them in simulation. These works range from comparing different 
 control algorithms in simulation, to designing novel controllers by combining existing controller algorithms.

Work done by \cite{prasad_tyagi_gupta_2014} analyses the performance of a PID and a LQR controller designed for a non linear inverted pendulum system. The controllers are tested in the presence of an external disturbance modelled as white noise. The tuning for the PID controller is done by trial and error and the LQR controller is designed for a given set of system dynamic parameters.

The work concluded the comparative advantage of LQR controller over PID controller, as the LQR controller does not require trial and error to tune the controller gains. The work also concluded that the LQR controller was more effective and robust than the PID controller, as the LQR controller was able to stabilise the pendulum at the unstable fixed point more quickly and more smoothly than the PID controller.

Work done by \cite{saraf_gupta_alivelu_parimi} analyses the performance of a PID and a LQR controller designed for a Quadrotor. The performance of the controllers is evaluated using Simulink. 

In order to design a suitable PID controller for the Quadrotor, the gains of 6 PID controllers had to be tuned for controlling the 6 different possible output states of the Quadrotor. This proved to be tedious, especially since the tuning is done using a trial and error method. In comparison, only two matrices had to be tuned to design a suitable LQR controller.

The work concluded that both the PID and LQR controller had similar response with respect to the settling time required to achieve the desired state response. However, the LQR controller performed the control more robustly as the number of overshoot peaks in the response of LQR controller were fewer than that in the response of the PID controller. Hence, the LQR controller's response was less aggressive.

Both the work make use of platforms such as Simulink to test the designed controllers. Much less work has been done to compare the system response of the controllers obtained from simulation to that obtained from a hardware setup with the same dynamics. This report focuses on making this comparison and seeks to validate the use of simulation for modelling the system response to controllers.

\section{Background}
\subsection{Robot dynamical system}
Robots in physical world can be generalised as a second-order dynamical system. Mathematically, this can be expressed as
\begin{equation}
\label{eqn:GE}
\ddot{q} = f(q, \dot{q}, u, t)
\end{equation}
where ${u}$ is the control vector.

Generally, the dynamics of the robot are affine in ${u}$. Hence the equation can be modified as below \cite{underactuated}
\begin{equation}
\label{eqn:GEM}
\ddot{q} = f_{1}(q, \dot{q}, t) + f_{2}(q, \dot{q}, t)u
\end{equation}

In the case of linear time-invariant systems, the above equation can be simplified into two first order differential equations, that can be represented using matrices as below \cite{state}

\begin{equation}
\label{eqn:GEST}
\dot{\mathbf{x}}=A\mathbf{x} + Bu
\end{equation}

The above equation is also known as the state transition equation of a dynamical system. The output of the dynamical system may be a function of both, the state variables and the control input. This can be expressed using matrices as below \cite{state}

\begin{equation}
\label{eqn:GECST}
\dot{\mathbf{y}}=C\mathbf{x} + Du
\end{equation}

\subsubsection{Pendulum system}
\begin{figure}[!htbp]
    \centering
    \includegraphics[width=60mm]{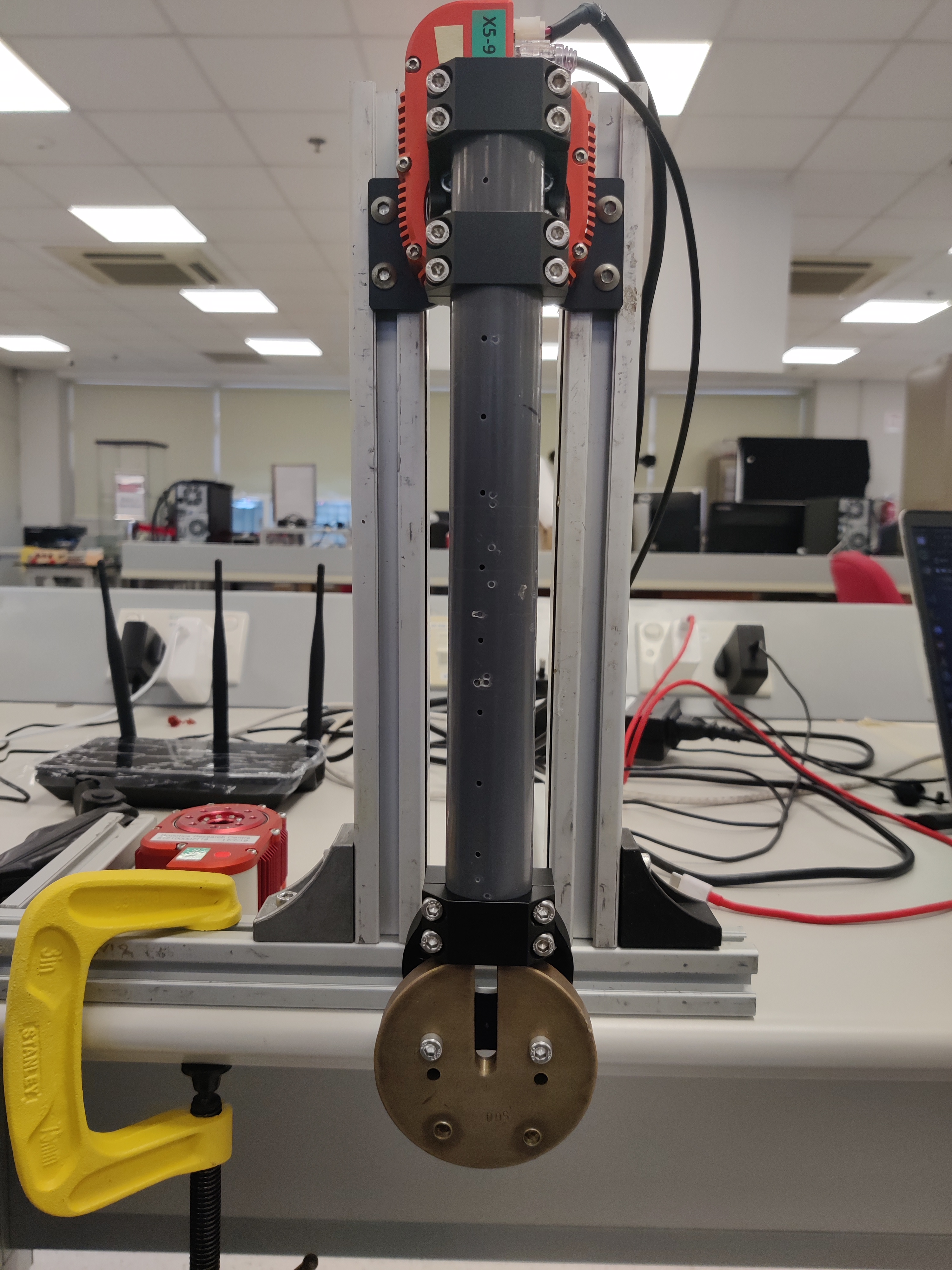}
    \caption{Pendulum system}
    \label{fig:PS}
\end{figure}

The Pendulum system \ref{fig:PS} is a mechanical system that consists of a mass attached to a link that can be rotated about an actuating rotary joint. If the link is mass less, then the system is known as a simple pendulum system. 

For the purposes of this report, the link in discussion, will be having mass along with another mass attached to its free end. Hence, the system is a generic pendulum system.

\FloatBarrier

The parameters of the Pendulum system are as described below:
\begin{itemize}
    \item ${m}$: Total mass of the point mass and the link, alongside other components making up the system
    \item ${l_c}$: Length of the link from the point of rotation to the center of mass of the system
    \item ${\theta}$: Angle made by the link with the vertical, reference being the point mass resting vertically down
    \item ${I_c}$: Rotational inertia of the system at the center of mass about the axis of rotation 
    \item ${u}$: Control input to the pendulum system
    \item ${b}$: Damping constant of the pendulum system
    \item ${g}$: Acceleration due to gravity
\end{itemize}

Since the system is a planar 1 degree of freedom system, the parameters ${m}$ and ${I_c}$ are scalar quantities.

The state of the Pendulum system at any time ${t}$ can be described using the vector, 
\begin{center}
   ${\mathbf{x} = \begin{bmatrix}
    \theta & \dot{\theta}
    \end{bmatrix}}$ 
\end{center}

The Equation of Motion of this system can be derived using the Lagrange method.

The energy of the system is given by:
\begin{equation}
\label{eqn:KE}
T=\frac{1}{2} m l_c^2 \dot{\theta^2} +  \frac{1}{2} I_c \dot{\theta^2}
\end{equation}

\begin{equation}
\label{eqn:PE}
U=-m g l_c \cos \theta
\end{equation}

where ${T}$ represents the Kinetic energy of the system and ${U}$ represents the Potential energy of the system.

The Lagrangian is defined as the following 
${L = T - U}$

Using Lagrangian mechanics, the equations of motion of a system can be derived.
\begin{equation}
\label{eqn:LEqn}
\frac{d}{dt} \frac{\partial L}{\partial {\dot{\theta}}} - \frac{\partial L}{\partial \theta} = Q
\end{equation}

where ${Q}$ represents the generalised forces acting on the system.

Using the Lagrangian mechanics stated above, the equations of motion of the pendulum system are derived as below
\begin{equation}
\label{eqn:PSEqn}
(m l_c^2 + I_c) \ddot{\theta}(t) + m g l_c \sin \theta(t) = Q  
\end{equation}

The generalized force ${Q}$ considered in the experimental setting can be defined to consist of a damping torque and a control input torque from the actuated joint. Hence, ${Q}$ can be modeled as below
\begin{equation}
\label{eqn:CI}
Q = -b \dot{\theta}(t) + u(t)  
\end{equation}

Hence, the dynamics of the pendulum system can be represented as 
\begin{equation}
\label{eqn:PSDEqn}
(m l_c^2 + I_c) \ddot{\theta}(t) + m g l_c \sin \theta(t) + b \dot{\theta}(t)= u(t)
\end{equation}

The above equations of motion can be rewritten in the standard manipulator equation form as:
\begin{equation}
\label{eqn:PSManipEqn}
M(q)\ddot{q} + C(q, \dot{q})\dot{q} = \tau_{g}(q) + Bu
\end{equation}

where the manipulator constants for pendulum system are as follows:
\begin{center}
    ${M(q)=m l_c^2 + I_c}$
\end{center} 
\begin{center}
    ${C(q, \dot{q})=b}$ 
\end{center}
\begin{center}
    ${\tau_{g}(q)=-m g l_c \sin \theta }$ 
\end{center}
\begin{center}
    ${B=1}$
\end{center}

The acceleration can then be evaluated as follows
\begin{equation}
\label{eqn:PSAccEqn}
\ddot{q}=M^{-1}[\tau_{g}(q) + Bu - C(q, \dot{q})\dot{q}]
\end{equation}

\subsubsection{Fixed points of a Dynamical System: Pendulum}
In the case of an over-damped pendulum the dynamics can be approximated as below 
\begin{equation}
\label{eqn:DPSEqn}
b\dot{x} + m g l\sin x = u_{0}
\end{equation}
Below is the plot of ${\dot{x}}$ vs ${x}$ 
\begin{figure}[!htbp]
    \centering
    \includegraphics[width=90mm]{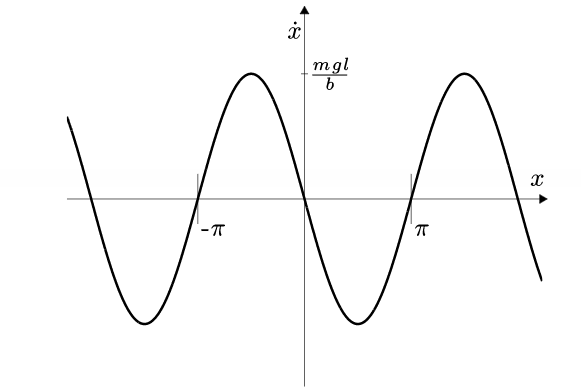}
    \caption{Overdamped pendulum state space plot}
    \label{fig:OVRDPS}
\end{figure}
\FloatBarrier

The points where ${\dot{x}=0}$ are known as the \textbf{steady state} or \textbf{fixed points} of the dynamical system. If the system is in one of these states, then it will continue to remain in that state unless disturbed.

The stability of a system at a given fixed point can be analysed by examining the response of the system in the vicinity of that fixed point. The points where ${\dot{x}>0}$, correspond to locations where the system tends to move away from the fixed points. Such fixed points are known as \textbf{unstable fixed points}. Correspondingly the points where ${\dot{x}<0}$ are known as \textbf{locally stable fixed points}.

Let ${\epsilon}$ denote an arbitrary small scalar quantity \cite{underactuated}.
\begin{itemize}
    \item Stability in the sense of Lyapunov (i.s.L). A dynamical system is said to be i.s.L around a fixed point ${x^*}$, if for every ${\epsilon>0}$, there exists a ${\delta>0}$ such that if ${||x(0)-x^*||<\delta}$, then for all ${t}$, ${||x(0)-x^*||<\epsilon}$
    \item A fixed point is unstable, if it is not locally stable i.s.L
\end{itemize}

The pendulum system has a i.s.L stable fixed point at state ${\mathbf{x}=\begin{bmatrix} \theta & \dot{\theta}
\end{bmatrix} = \begin{bmatrix}
 0 & 0
\end{bmatrix}}$, and an unstable fixed point at the state ${\mathbf{x}=\begin{bmatrix} \theta & \dot{\theta}
\end{bmatrix} = \begin{bmatrix}
 \pi & 0
\end{bmatrix}}$.

\subsubsection{Linear Dynamical Systems}
Any given non-linear dynamical system can be linearized about a point in its state space by evaluating the Taylor Series expansion of its non-linear dynamics about the point of interest. This is discussed in more detail in the subsequent sections. This enables representing a non-linear dynamical system as a linear dynamical system in the vicinity of the point of interest as below:
\begin{equation}
\label{eqn:stateTransition}
\dot{\mathbf{x}}=A\mathbf{x} + Bu
\end{equation}

The general solution to the given linear differential equation with zero initial condition is as follows
\begin{equation}
\label{eqn:stateTransitionSoln}
\mathbf{x}(t)=e^{At}B * u(t)
\end{equation}
The solution is the convolution of the control input ${u(t)}$ with the kernel ${e^{At}B}$

\subsubsection{Linearizing the manipulator equations}
Any dynamical system can be linearized about a point in its state space using Taylor series expansion.
Given a dynamical system represented by the equation below
\begin{equation}
\label{eqn:GDSEqn}
\dot{\mathbf{x}}=f(\mathbf{x}, u)
\end{equation}

The Taylor expansion of the above dynamical system around a fixed point ${f(\mathbf{x}^*, u^*)}$ can be represented by the below equation
\begin{equation}
\label{eqn:TaylorEqn}
\dot{\mathbf{x}}=f(\mathbf{x}, u)=f(\mathbf{x}^*, u^*)+\left[\frac{\partial f}{\partial \mathbf{x}}\right]_{\mathbf{x}=\mathbf{x}^*, u=u^*}(\mathbf{x}-\mathbf{x}^*) + \left[\frac{\partial f}{\partial u}\right]_{\mathbf{x}=\mathbf{x}^*, u=u^*}(u-u^*)
\end{equation}
If the point we are linearizing about is a fixed point, then ${f(\mathbf{x}^*, u^*)}$ is zero. The linearized equation can be represented as 
\begin{equation}
\label{eqn:TaylorLinEqn}
\dot{\mathbf{x}}=\begin{bmatrix}
\dot{q}\\
M^{-1}(q)[\tau_{g}(q)+B(q)u-C(q, \dot{q})\dot{q}]
\end{bmatrix} 
\approx A_{\text{lin}}(\mathbf{x}-\mathbf{x}^*) + B_{\text{lin}}(u-u^*)
\end{equation}
${A_{\text{lin}}}$ and ${B_{\text{lin}}}$ are constant matrices. Defining ${\Bar{\mathbf{x}}=(\mathbf{x}-\mathbf{x}^*)}$ and ${\Bar{u}=(u-u^*)}$, we can rewrite the linearized equation as follows
\begin{equation}
\label{eqn:SimpTaylorLinEqn}
\dot{\Bar{\mathbf{x}}}=A_{\text{lin}}\Bar{\mathbf{x}} + B_{\text{lin}}\Bar{u}
\end{equation}

Solving for the constant matrices ${A_{\text{lin}}}$ and ${B_{\text{lin}}}$ in the general manipulator equation about a fixed point, we arrive at the following results \cite{underactuated}.
\begin{equation}
\label{eqn:Alin}
A_{\text{lin}}=\begin{bmatrix}
\mathbf{0} & \mathbf{I}\\
M^{-1}\frac{\partial \tau_{g}}{\partial q}+\sum_{j}M^{-1}\frac{\partial B_{j}}{\partial q}u_{j} & \mathbf{0}
\end{bmatrix}_{\mathbf{x}=\mathbf{x}^*, u=u^*}
\end{equation}

\begin{equation}
\label{eqn:Blin}
B_{\text{lin}}=\begin{bmatrix}
\mathbf{0} \\
M^{-1}B
\end{bmatrix}_{\mathbf{x}=\mathbf{x}^*, u=u^*}
\end{equation}

\subsection{Optimal control techniques to balance a pendulum system}

\subsubsection{Linear Quadratic Regulator}
Linear Quadratic Regulator is a control technique to derive an optimal ${{K}_{r}}$ (controller gain matrix), given a set of constraints or cost on the parameters of the dynamical system. This assumes that the dynamical system is controllable and observable. A dynamical system is said to be observable if it is possible to build a full state estimate of the system from the sensor measurements. In the trivial case this implies that for the below equation
\begin{equation}
\mathbf{y} = C\mathbf{x} + D
\end{equation}
${C=I}$ and ${D=0}$, which implies that the output of the system, ${\mathbf{y}}$ is equal to the current state of the system, ${\mathbf{x}}$

A dynamical system can be made stable by placing the eigenvalues in the left-half of the complex plane. It is possible to place these poles as far left in the complex plane as required, but this results in a number of disadvantages that degrade the performance of the system. These disadvantages are listed below \cite{brunton_2019_datadriven}.
\begin{itemize}
    \item Placing a pole in the far left of the complex plane, results in a system whose eigenvalues have high magnitudes. This results in expensive control expenditure.
    \item As a consequence of expensive control expenditure, the system may exceed the maximum allowable value of the control variable that the hardware can provide.
    \item Over stabilisation of a system may cause it to respond aggressively to noise and disturbances, especially those with high frequency, and can cause the system to jitter.
\end{itemize}

Hence, there arises a need to strike a balance between 
\begin{itemize}
    \item Stability of the system
    \item Constrained control expenditure
\end{itemize}

This is the goal of optimal control. 

In order to quantify the balance between the above two mentioned control objectives, a cost function is developed as below
\begin{equation}
\label{eqn:LQRCost}
J(t) = \int_{0}^{t} \mathbf{x}(\tau)\mathbf{Q}\mathbf{x}(\tau) + \mathbf{u}(\tau)\mathbf{R}\mathbf{u}(\tau)\, d\tau
\end{equation}
The cost function regulates the two control objectives. The cost matrices ${\mathbf{Q}}$ and ${\mathbf{R}}$ weigh the cost of deviations from the regulated state and from regulated actuation respectively. Matrix  ${\mathbf{Q}}$ is positive semi-definite and 
${\mathbf{R}}$ is positive-definite. The matrices are diagonal matrices and the elements of the diagonal can be altered to alter the relative cost given to each basis vector in the matrix.

This transforms into a well-posed optimization problem. The technique is named LQR as it attempts to minimise a Quadratic cost function for a linearised dynamical system using a linear control law.

The analytical solution for the optimal controller gain, ${{K}_{r}}$, for the above quadratic cost function is given by
\begin{equation}
\label{eqn:LQRSoln}
{K}_{r} = R^{-1}BX
\end{equation}

where ${X}$ is the solution to an algebraic Riccati equation
\begin{equation}
\label{eqn:RiccatiEqn}
AX + XA - XBR^{-1}BX + Q = 0
\end{equation}

The control law, ${u = {K}_{r}\mathbf{x}}$, is the optimal solution to the quadratic cost function, subject to the dynamical constraint of the linearised dynamical system.

\section{Experimental determination of
parameters of system dynamics}
In order to develop an approximated system model to simulate, the parameters of the system dynamics have to be determined. This chapter focuses on the experimental setup, which includes the hardware used and the experiments used in order to obtain the required parameters.

The dynamical equation for the pendulum system as given in \ref{eqn:PSDEqn} is 
\begin{center}
   ${(m l_c^2 + I_c) \ddot{\theta}(t) + m g l_c \sin \theta(t) + b \dot{\theta}(t)= u(t)}$ 
\end{center}

The above equation can be abstracted in the standard form as below 
\begin{equation}
    \label{eqn:PSDEqnConstants}
    M_c\ddot{\theta} + b_c\dot{\theta} + G_c \sin \theta = u
\end{equation}

It is evident that the parameters ${M_c, b_c}$ and ${G_{c}}$ are constants.

The parameters to be determined from the equation are as given below:
\begin{itemize}
        \item Inertial constant ${M_c}$
    \item Damping constant ${b_c}$
    \item Gravitational force coefficient ${G_{c}}$
\end{itemize}

This chapter discusses the experiments used in order to determine these parameters.

\subsection{Hardware components used}
The hardware setup for all subsequent experiments consists of a subset of the below mentioned components. 

\begin{itemize}
    \item HEBI X5-9 actuator and power cable
    \item Hollow pipe (link of the simple pendulum system)
    \item A-2089-02 HEBI bracket set
    \item Cylindrical mass (mass bob of the simple pendulum system)
    \item M5 screws
    \item Router
    \item LAN cable
    \item Aluminium Extrusion
\end{itemize}

\begin{figure}[!htbp]
    \centering
    \includegraphics[width=110mm]{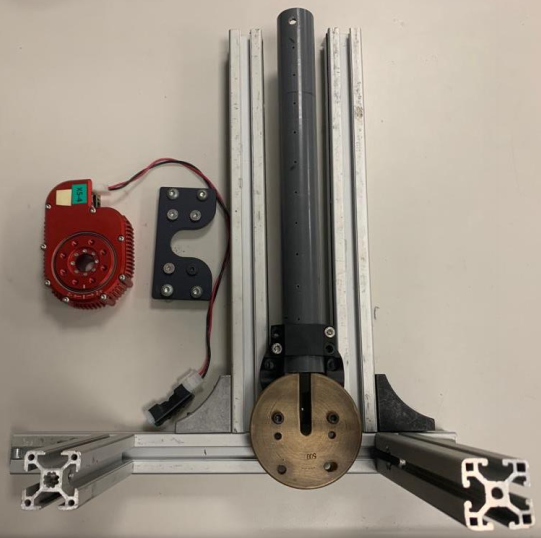}
    \caption{Experiment hardware components}
    \label{fig:HEBIHW}
\end{figure}
\FloatBarrier

\subsection{HEBI actuator}
HEBI X-series actuators are smart series-elastic actuators consisting of a brushless motor, gear-train, spring, encoder and the required control electronics, all integrated into a compact package, that can be run on DC voltages and uses 10/100 Mbps Ethernet for communication \cite{hebi}.

The actuators are capable of providing simultaneous control of position, velocity and torque using standard PID controllers. The actuator provides four different control strategies which are a result of the combination of the above mentioned PID controllers in different configurations. 

HEBI Robotics provides Scope GUI that can be used to interact with the actuator. They also provide Python, C++ and MATLAB code API's as well. Scope GUI and the Python code API have been used for the experiments. The below quantities are recorded during all subsequent experiments.

\begin{table}[h!]
\centering
\begin{tabular}{|c|} 
 \hline
 Quantity\\ [0.5ex] 
 \hline \hline
 Duration\\ 
 \hline
 Position feedback \\
 \hline
 Velocity feedback \\
 \hline
 Effort feedback \\
 \hline
 Position commanded \\ 
 \hline
 Velocity commanded \\
 \hline
 Effort commanded \\
 \hline
 PWM commanded \\
 \hline
 Motor current \\
 \hline
 Motor winding current \\
 \hline
\end{tabular}
\caption{Quantities recorded during experiments}
\label{table:Recorded experimental quantities}
\end{table}

\begin{figure}[!htbp]
    \centering
    \includegraphics[width=110mm]{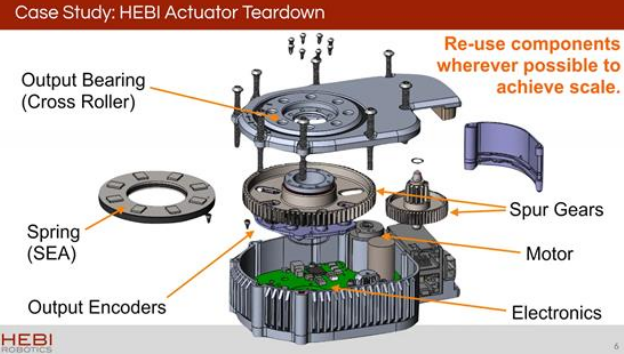}
    \caption{HEBI Actuator Teardown}
    \label{fig:HEBITD}
\end{figure}
\FloatBarrier

The section below will explore in more detail, the HEBI X5-9 actuator and the two control strategies which are used subsequently in this report.

\subsubsection{HEBI X5-9 specifications}
\begin{figure}[!htbp]
    \centering
    \includegraphics[width=120mm]{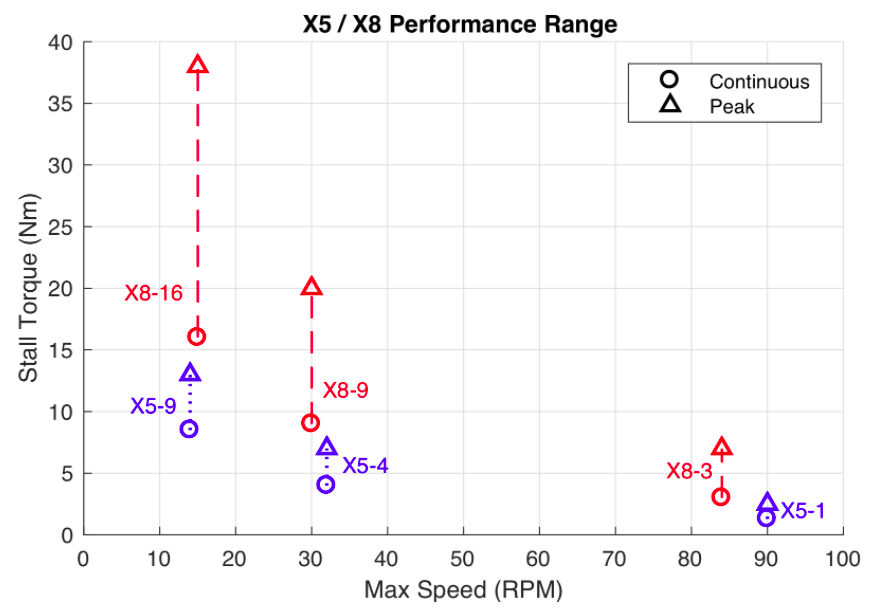}
    \caption{HEBI X-series Performance}
    \label{fig:HEBIPF}
\end{figure}

The stall torque and maximum speed performance of the X5-9 actuator is shown in \ref{fig:HEBIPF}

\FloatBarrier

\begin{figure}[!htbp]
    \centering
    \includegraphics[width=120mm]{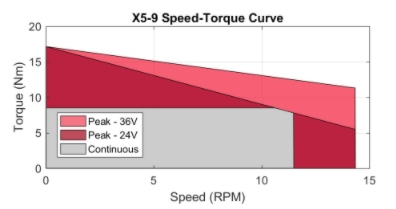}
    \caption{X5-9-speed-torque curve}
    \label{fig:X5-9ST}
\end{figure}

The speed, torque curve of the X5-9 actuator is shown in \ref{fig:X5-9ST}

\FloatBarrier

\begin{figure}[!htbp]
    \centering
    \includegraphics[width=140mm]{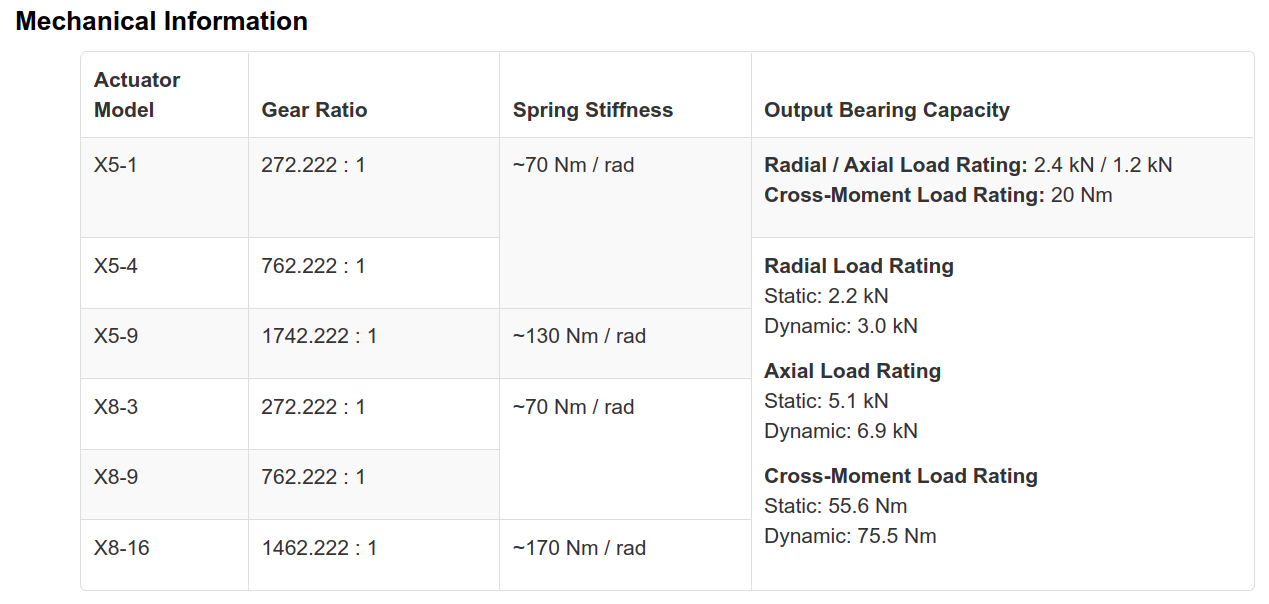}
    \caption{HEBI Actuator Mechanical details}
    \label{fig:HEBIMech}
\end{figure}

The electrical and mechanical information of the X5-9 actuator is shown in \ref{fig:HEBIMech}

\FloatBarrier

\begin{figure}[!htbp]    
    \centering
    \includegraphics[width=140mm]{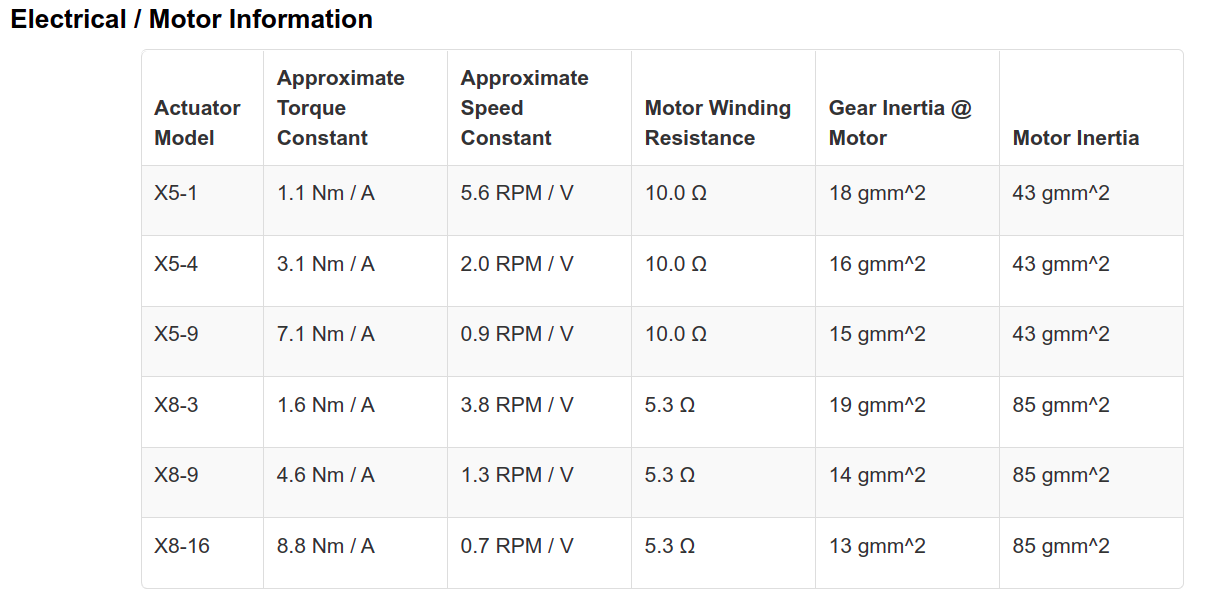}
    \caption{HEBI Actuator Electrical details}
    \label{fig:HEBIElec}
\end{figure}
\FloatBarrier

\subsubsection{Control strategies}
Out of the four available control strategies in HEBI X5-9 actuator, two strategies namely, Strategy 1 and Strategy 3 are discussed here. These control strategies are used subsequently in the report.

\textbf{Strategy 1 / Direct PWM} \newline
The commanded torque directly sets the motor PWM, from -1 to +1. This strategy does not use feedback from any of the sensors and essentially functions as an open loop feed-forward system.

\begin{figure}[!htbp]
    \centering
    \includegraphics[width=120mm]{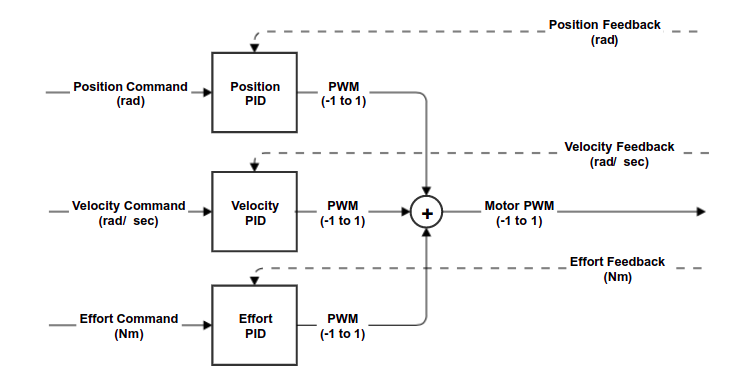}
    \caption{HEBI Actuator Strategy 3}
    \label{fig:HEBIStrategy3}
\end{figure}

\textbf{Strategy 3} \newline
In Strategy 3, the position, velocity and effort (torque) controllers generate individual PWM commands, which is summed to generate a final PWM output command between -1 and +1. \ref{fig:HEBIStrategy3} illustrates this control strategy

\FloatBarrier

The equations governing the internal working of a motor is as below
\begin{figure}[!htbp]
    \centering
    \includegraphics[width=120mm]{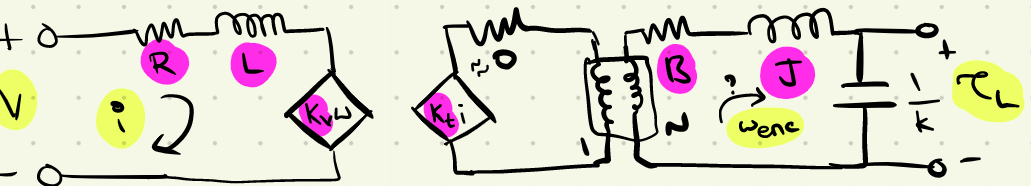}
    \caption{Motor Equations}
    \label{fig:MotorEqn}
\end{figure}
\FloatBarrier

\begin{equation}
    \label{eqn:MotorEqn1}
    V - Ri - L\frac{di}{dt} - k_v\dot{\theta} = 0
\end{equation}

\begin{equation}
    \label{eqn:MotorEqn2}
    {\tau}_L - J\ddot{\theta} - B\dot{\theta}= 0
\end{equation}

\subsection{Determination of Inertial and Damping constant}
This experiment details the procedure to find out the damping and the inertial constant of the hardware pendulum system setup. The damping constant due to air is insignificant in comparison to the damping constant due to the X5-9 actuator and system. 

The experimental setup is as shown in the figure below

\begin{figure}[!htbp]
    \centering
    \includegraphics[width=110mm]{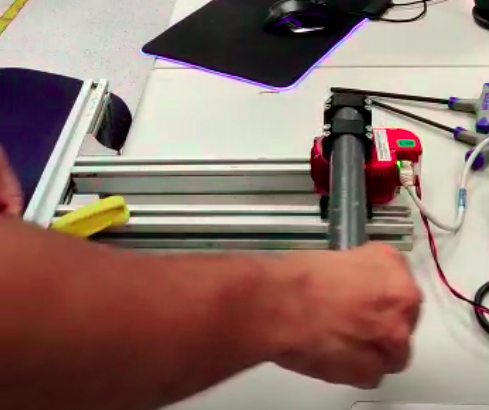}
    \caption{Inertial and Damping constant experiment setup}
    \label{fig:HorSetup}
\end{figure}
\FloatBarrier

The experimental procedure is described below
\begin{itemize}
    \item The experimental setup consists of the X5-9 actuator, aluminium extrusions, hollow pipe, cylindrical mass and the A-2089-02 HEBI bracket set.
    \item The X5-9 actuator is setup and connected to a PC as per details provided on the HEBI robotics website. The pendulum system is setup horizontally flat on the table. This voids the contribution from gravity constant and isolates the contribution from parameters ${M_c}$ and ${b_c}$.
    \item Strategy 1 / Direct PWM control strategy is set on the actuator. 
    \item The pendulum is rotated by applying a torque using hand and slowly accelerating it from rest. 
    \item The pendulum is rotated from a start angle of ${0 rad}$ up to ${\pi rad}$, and then brought back to the initial state. The same step is repeated in the other direction as well. 
    \item During the motion, a ${PWM_{cmd} = 0}$ is commanded continuously to the actuator and the quantities mentioned in table \ref{table:Recorded experimental quantities} are recorded. This voids any contribution from the electrical side of the motor as per \ref{eqn:MotorEqn1}.
    \item Using the data from the Duration and Velocity Feedback, the acceleration experienced by the system can be calculated for each time step using the below relation
    \begin{equation}
        a_{fbk} = \frac{v_{fbk, 2} - v_{fbk, 1}}{t_{2} - t_{1}}
    \end{equation}
    \item The experiment is repeated for 6 trials, and the data recorded is plotted 
\end{itemize}
    
\begin{figure}[!htbp]
    \centering
    \includegraphics[width=150mm]{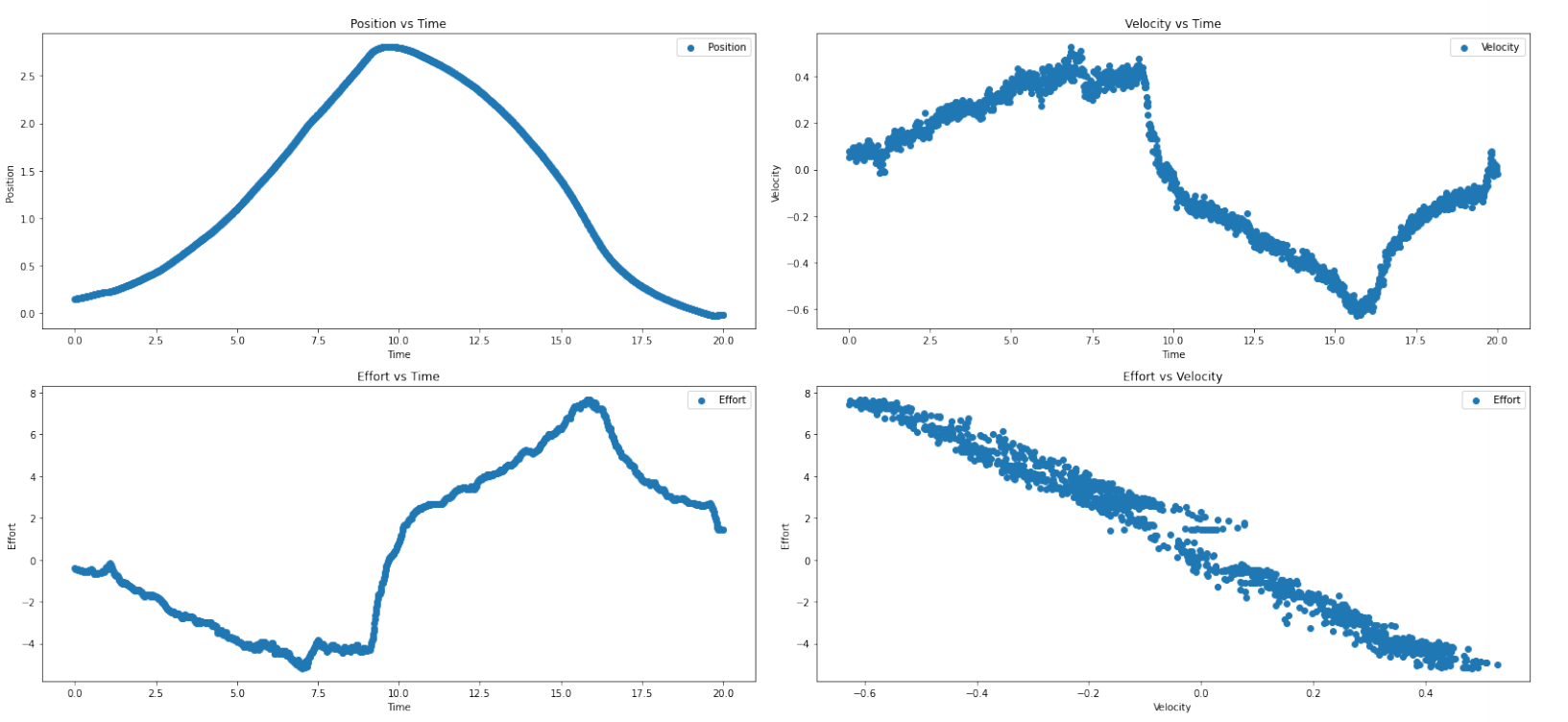}
    \caption{Inertia and Damping Exp 1}
    \label{fig:ID1}
\end{figure}
\FloatBarrier
\begin{figure}[!htbp]
    \centering
    \includegraphics[width=150mm]{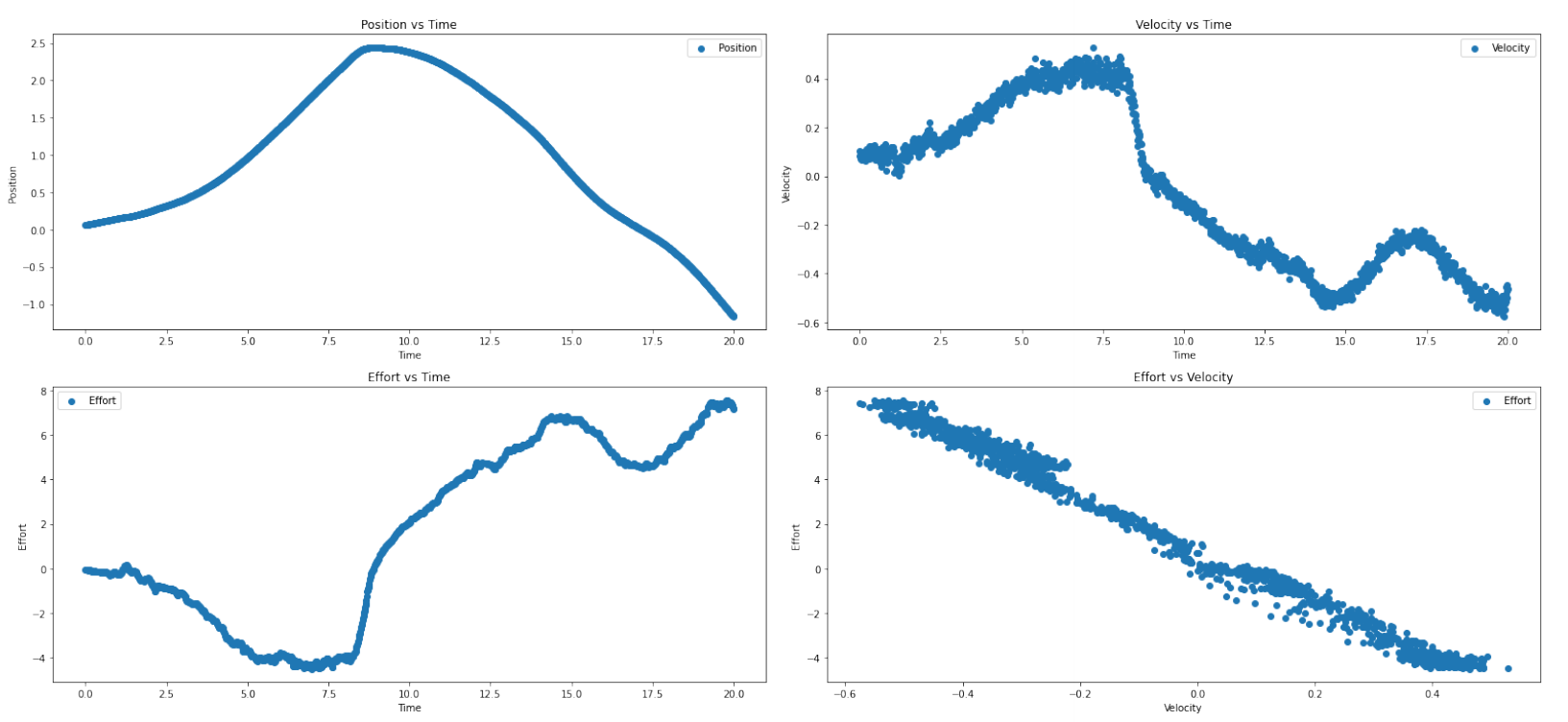}
    \caption{Inertia and Damping Exp 2}
    \label{fig:ID2}
\end{figure}
\FloatBarrier
\begin{figure}[!htbp]
    \centering
    \includegraphics[width=150mm]{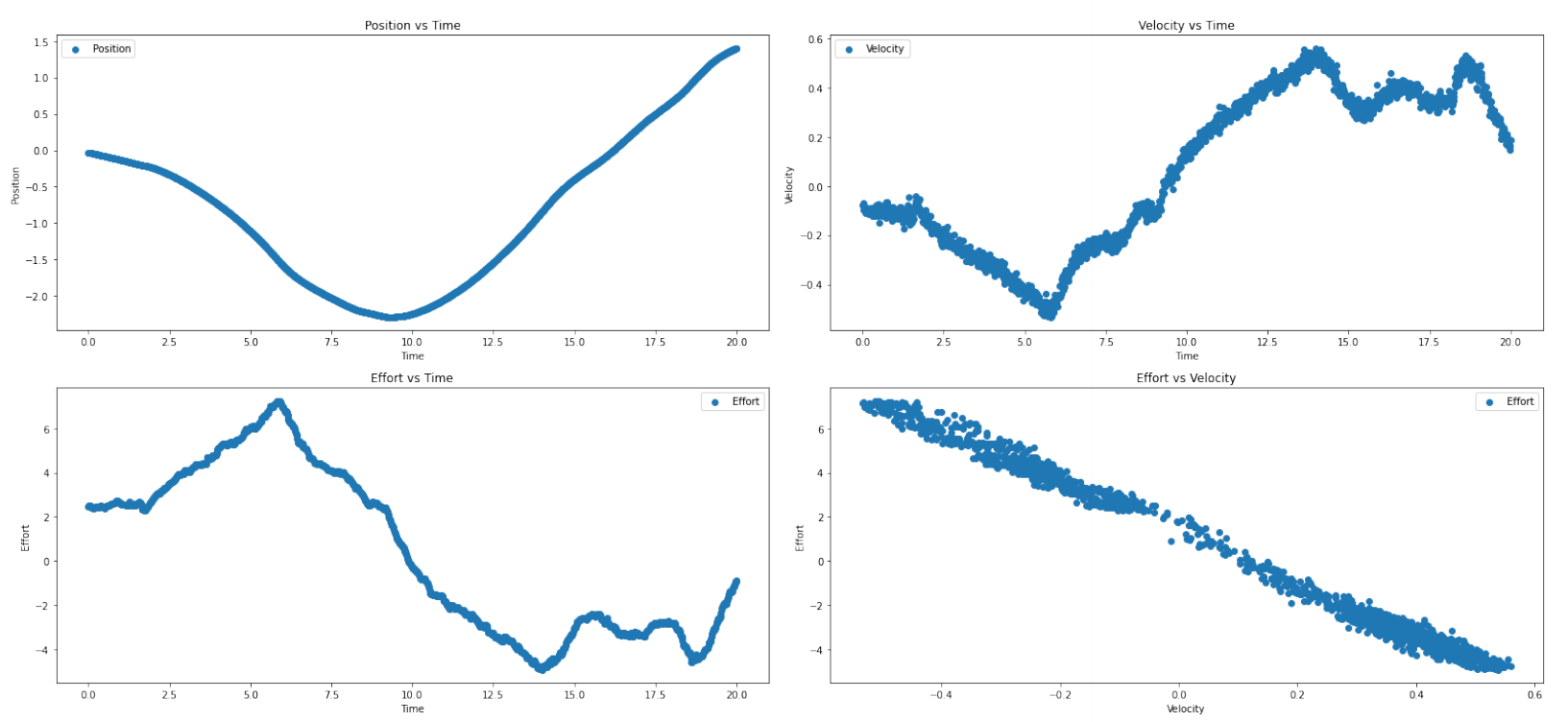}
    \caption{Inertia and Damping Exp 3}
    \label{fig:ID3}
\end{figure}
\FloatBarrier
\begin{figure}[!htbp]
    \centering
    \includegraphics[width=150mm]{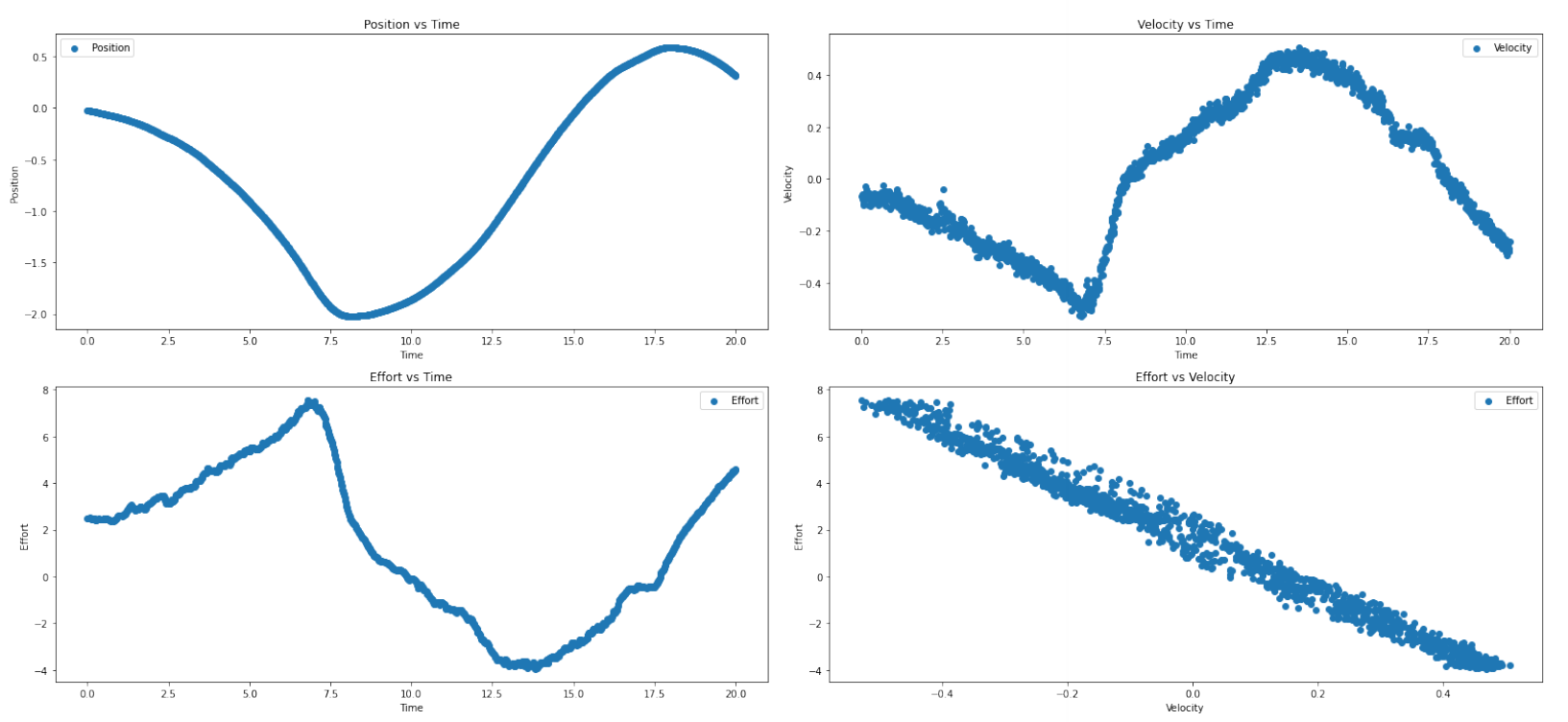}
    \caption{Inertia and Damping Exp 4}
    \label{fig:ID4}
\end{figure}
\FloatBarrier
\begin{figure}[!htbp]
    \centering
    \includegraphics[width=150mm]{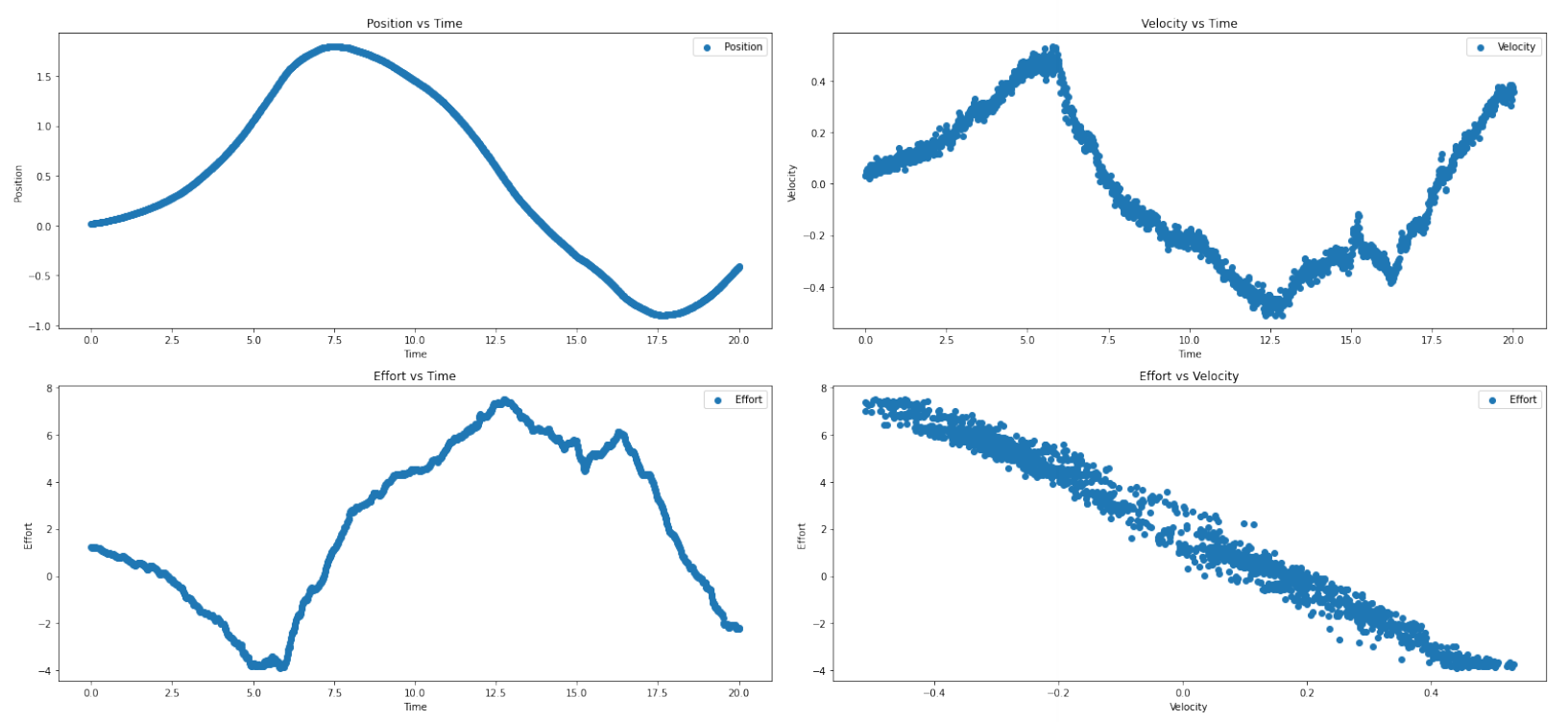}
    \caption{Inertia and Damping Exp 5}
    \label{fig:ID5}
\end{figure}
\FloatBarrier
\begin{figure}[!htbp]
    \centering
    \includegraphics[width=150mm]{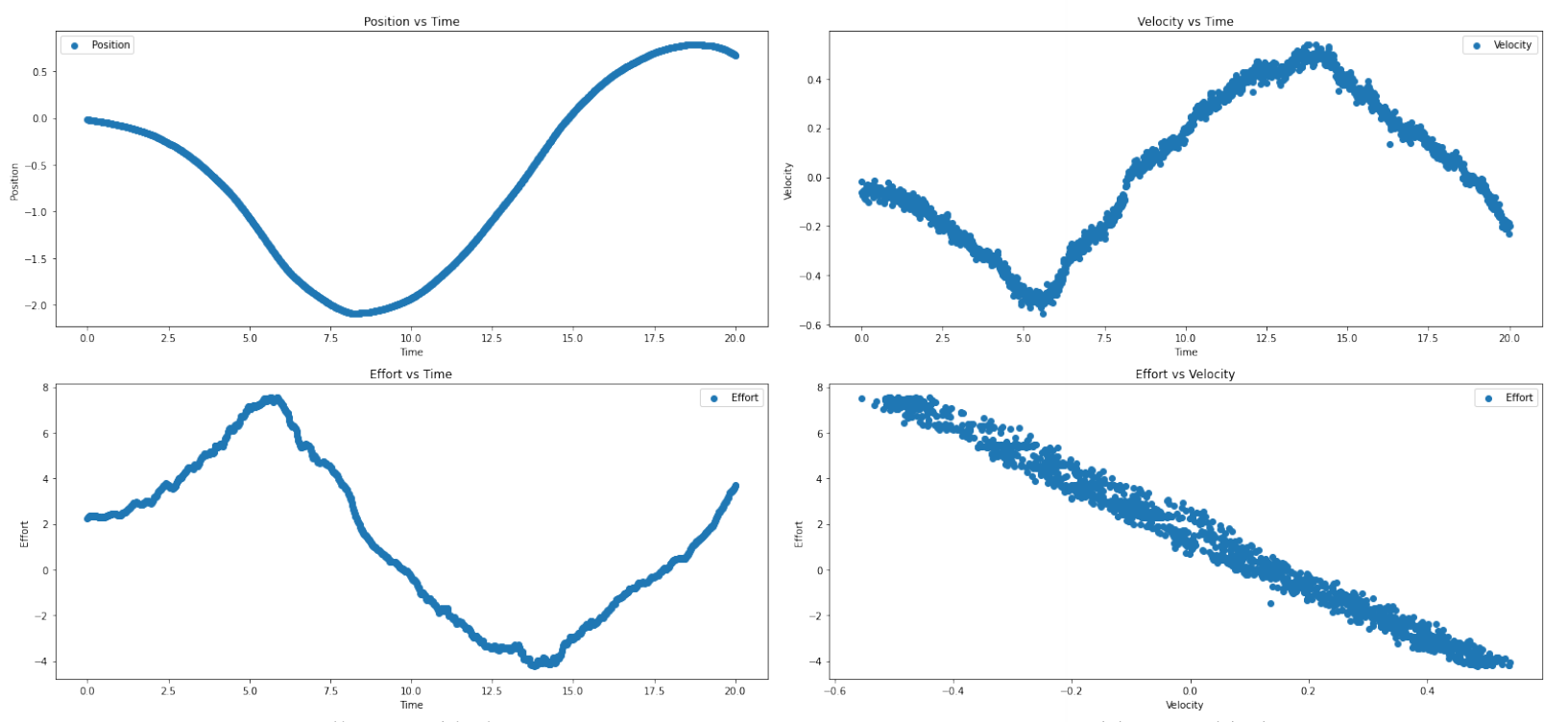}
    \caption{Inertia and Damping Exp 6}
    \label{fig:ID6}
\end{figure}
\FloatBarrier

From the Effort Feedback vs Velocity Feedback graphs, a linear relationship can be observed as expected from \ref{eqn:PSDEqn} and \ref{eqn:PSDEqnConstants}.

Using Effort Feedback as the dependant variable and Velocity Feedback, Acceleration as the independent variable, a simple Least Squares Regression model can be setup for each of the experiment. The result of the regression model for each of the experiments is shown below. The Velocity coefficient corresponds to the damping constant of the system and the Acceleration coefficient corresponds to the Inertial constant of the system.

\begin{figure}[!htbp]
    \centering
    \includegraphics[width=140mm]{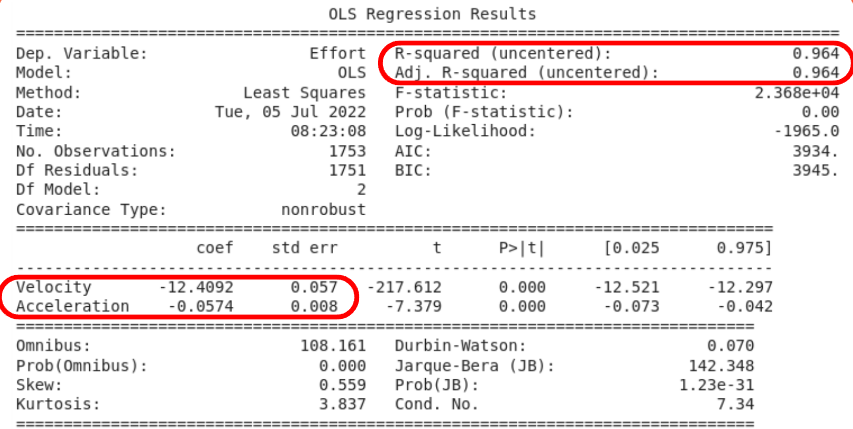}
    \caption{Inertia and Damping constant OLS}
    \label{fig:IDOLS}
\end{figure}
\FloatBarrier
\begin{figure}[!htbp]
    \centering
    \includegraphics[width=65mm]{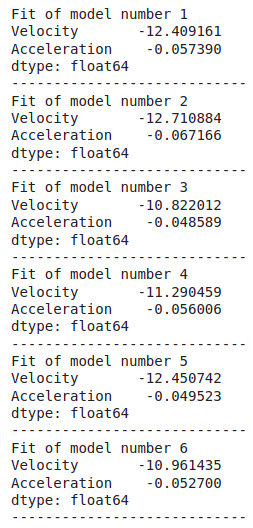}
    \caption{Inertia and Damping experiment regressed coefficients}
    \label{fig:IDCoeff}
\end{figure}
\FloatBarrier

The negative sign in the velocity coefficients obtained in the above experiments is due to HEBI motor recording an external torque with a negative magnitude.
An average of the two coefficients across all 6 trials is computed and the below values are obtained 

${\mathbf{b_c} : 11.77 \frac{Ns}{m}}$

${\mathbf{M_c} : 0.055 \frac{kg}{m^2}}$

\subsection{Determination of Gravitational force coefficient of the system}
This experiment details the procedure to find the gravitational force coefficient of the pendulum system. The experiment is premised on the concept of center of mass of a physical system. The components constituting the pendulum system can be simplified to a point mass located at a fixed distance from the actuated joint. This fixed distance is the length to the center of mass of the system. 

The experimental setup is similar to a vertical pendulum, as shown in the figure below

\begin{figure}[!htbp]
    \centering
    \includegraphics[width=90mm]{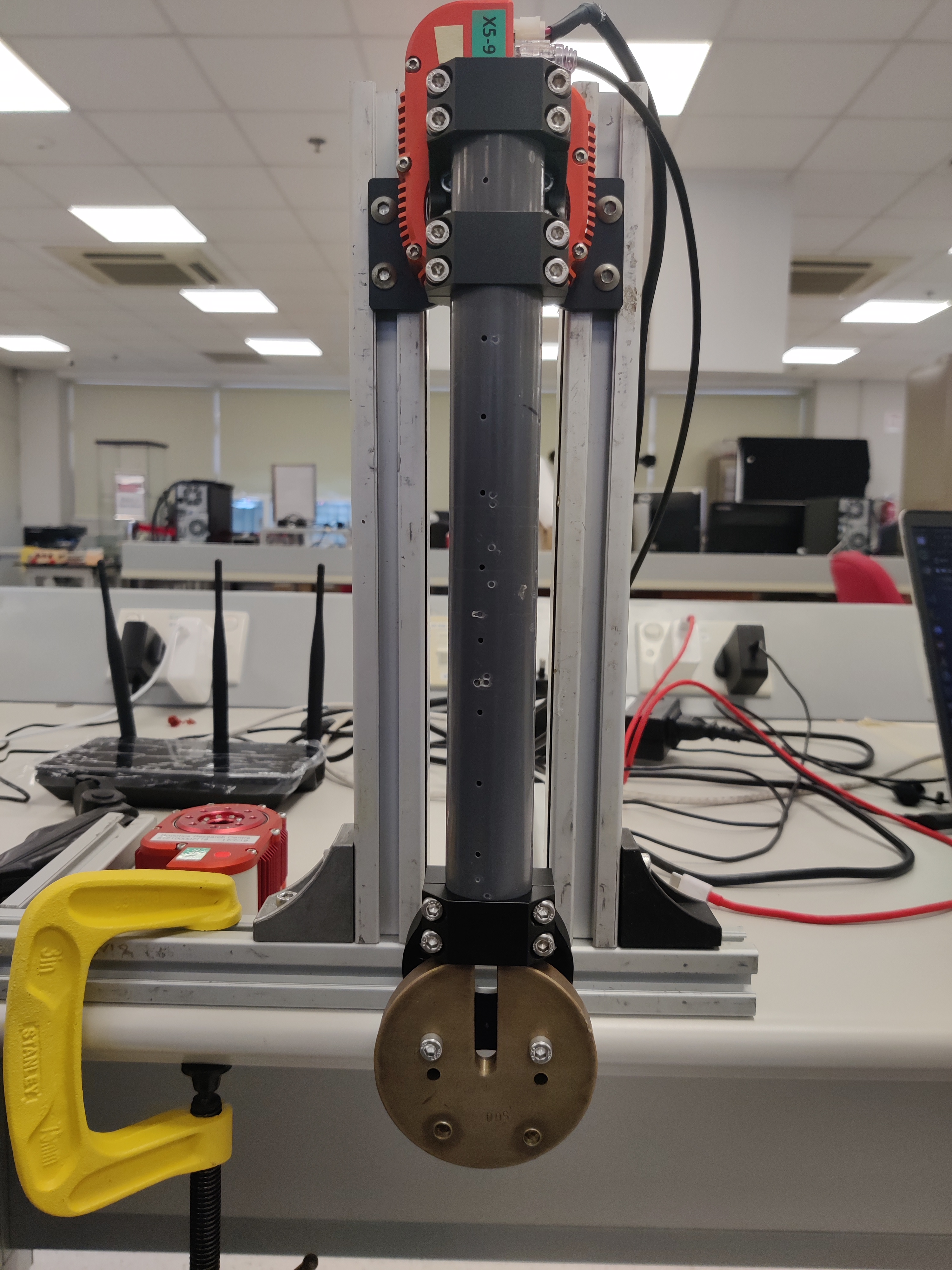}
    \caption{Gravitational coefficient experiment setup}
    \label{fig:VPSetup}
\end{figure}
\FloatBarrier

The experimental procedure is described below

\begin{itemize}
    \item The experimental setup consists of the X5-9 actuator, aluminium extrusions, hollow pipe, cylindrical mass and the A-2089-02 HEBI bracket set.
    \item The X5-9 actuator is setup and connected to the PC as per details provided on the HEBI robotics website. The pendulum system is setup vertically.
    \item Strategy 3 is set on the actuator with both position and velocity control to command smooth trajectory.
    \item Default gains as provided on the HEBI robotics website are used for both controls.
    \item The experiment consists of two phases: the \textit{static phase} and the \textit{dynamic phase}, which occur alternatively throughout the experiment.
    \item The pendulum is started at ${0 \degree}$, and then commanded to go to positions with increments of ${10 \degree}$. The motion of the pendulum from one position to another constitutes the dynamic phase.
    \item The commanded position vector is generated by interpolating a linear position vector between the start and end position with increments of ${0.002 rad}$ or ${0.1146 \degree}$. 
    \item The commanded velocity vector is generated as a half cycle sinusoidal curve having the same length as the commanded position vector and a maximum amplitude of ${0.3 rad/s}$ in order to offset the torque due to gravity.
    \item The trajectory is generated by commanding the position and velocity vector simultaneously with a command frequency of ${100Hz}$.
    \item During the motion to the desired position, the quantities mentioned in table ${T1}$ were recorded.
    \item After reaching the commanded position, the system is held in place by commanding a combination of the desired position and a velocity of ${0rad/s}$ to the controller. This constitutes the static phase of the experiment.
    \item During the static phase of the experiment, the quantities mentioned in \ref{table:Recorded experimental quantities} are recorded. During the static phase the contribution from parameters ${M_c}$ and ${b_c}$ are voided.
    \item One experiment consists of the positive half of the cycle (angles from ${0 \degree}$ to ${90 \degree}$ with increments of ${10 \degree}$, and brought back to ${0 \degree}$ from  ${90 \degree}$ with increments of ${-10 \degree}$) and negative half of the cycle (angles from ${0 \degree}$ to ${-90 \degree}$ with increments of ${-10 \degree}$, and brought back to ${0 \degree}$ from  ${-90 \degree}$ with increments of ${10 \degree}$). This consists of both the dynamic phase and the static phase.
    \item The experiment is repeated for 3 trials, and the recorded data is plotted as in the graphs below 
\end{itemize}

\begin{figure}[!htbp]
    \centering
    \includegraphics[width=140mm]{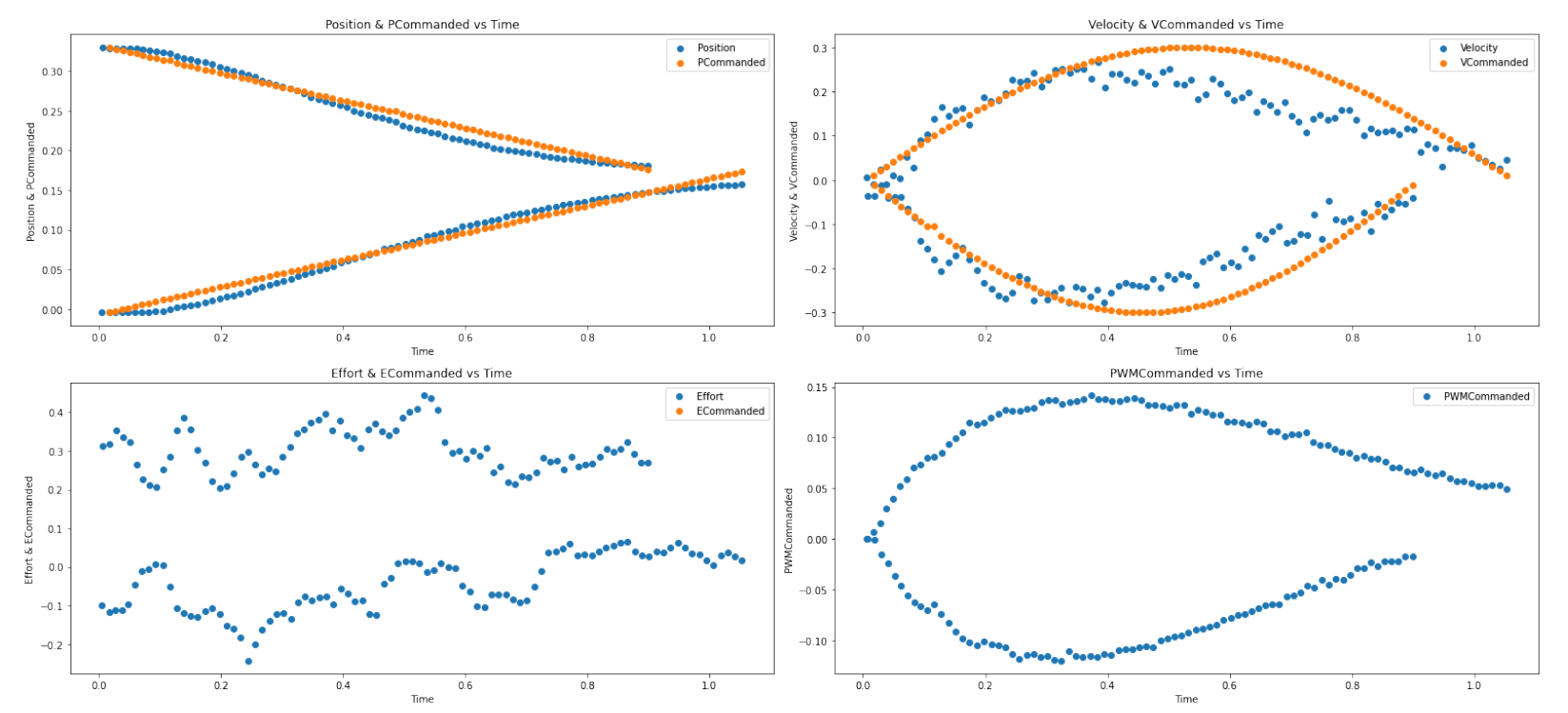}
    \caption{Dynamic Phase: Gravitational force coefficient plots for ${0 \degree}$ to ${10 \degree}$ and ${20 \degree}$ to ${10 \degree}$}
    \label{fig:E2DP01020}
\end{figure}
\FloatBarrier
\begin{figure}[!htbp]
    \centering
    \includegraphics[width=140mm]{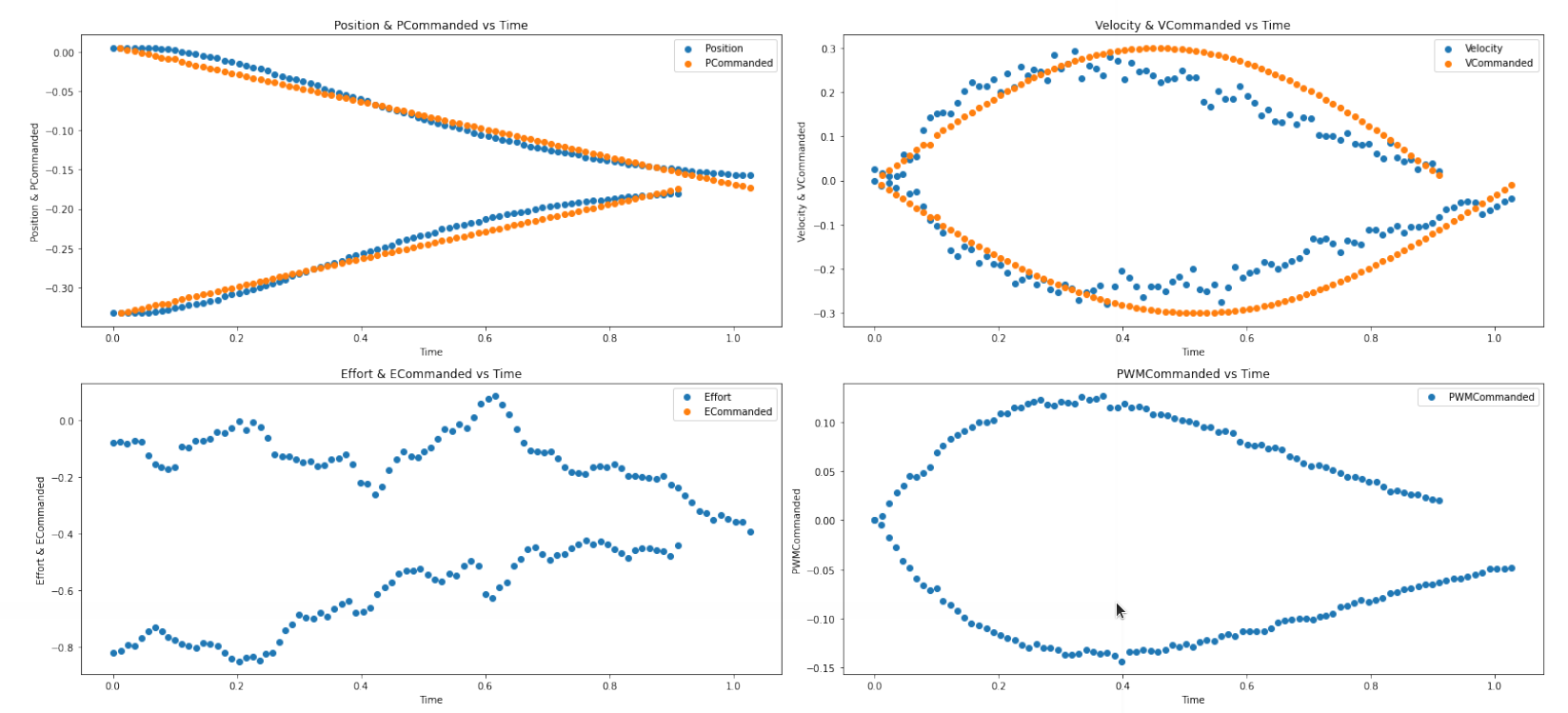}
    \caption{Dynamic Phase: Gravitational force coefficient plots for ${0 \degree}$ to ${-10 \degree}$ and ${-20 \degree}$ to ${-10 \degree}$}
    \label{fig:E2DP-01020}
\end{figure}
\FloatBarrier

\ref{fig:E2DP01020}, \ref{fig:E2DP-01020} are for the dynamic phase. \ref{fig:E2DP01020}, \ref{fig:E2DP-01020} show the position and velocity curve traced by the system when commanded to move from one angle to the other. It is evident from the ${PWM_{cmd}}$ vs Time graph, that the ${PWM_{cmd}}$ values are smooth, hence ensuring that the motor does not experience sudden jerks, which may cause inaccurate feedback readings due to factors such as Hysteresis. 

In \ref{fig:E2DP-01020}, the two different plots represent the feedback captured during the positive cycle of motion, which involves going from  ${0 \degree}$ to ${10 \degree}$, and during the negative cycle of motion, which involves going from  ${20 \degree}$ to ${10 \degree}$. The time axis for the plots start from 0, in order to make it easier to visualize the motion. Similar plotting strategy is used for the other plot.

\ref{fig:E2DP01020}, \ref{fig:E2DP-01020} show the dynamic phase involved in moving from one angle to another for two different starting angles in one experiment, out of the total three experiments. 

\begin{figure}[!htbp]
    \centering
    \includegraphics[width=130mm]{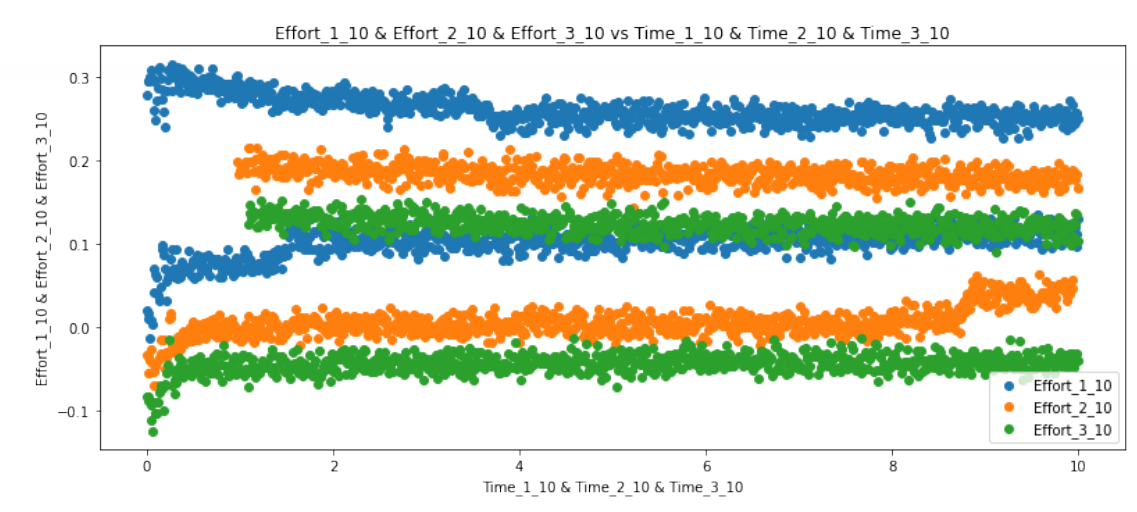}
    \caption{Static phase: Gravitational force coefficient Plots for ${10 \degree}$ }
    \label{fig:E2SP10}
\end{figure}
\FloatBarrier
\begin{figure}[!htbp]
    \centering
    \includegraphics[width=130mm]{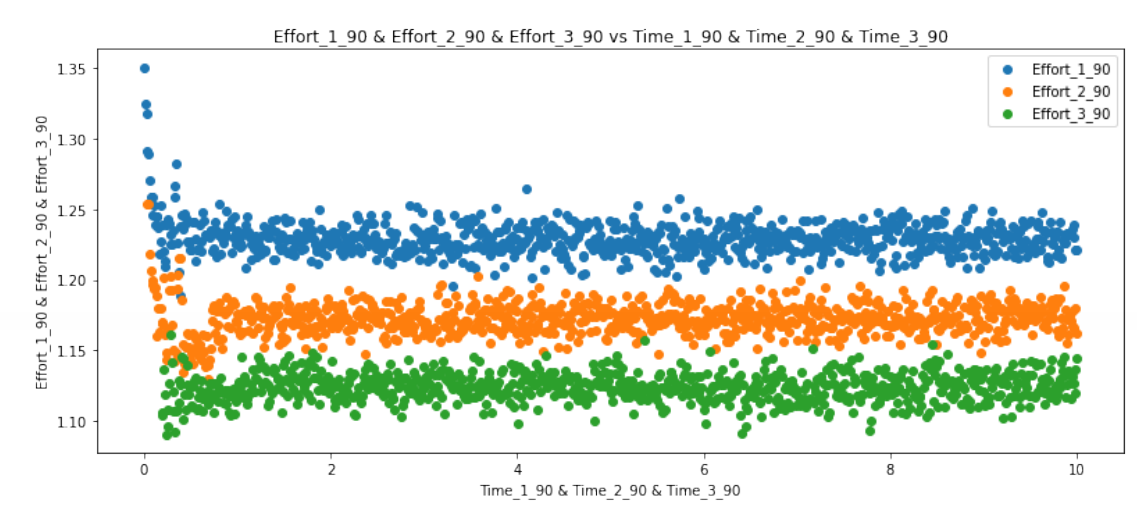}
    \caption{Static phase: Gravitational force coefficient Plots for ${90 \degree}$}
    \label{fig:E2SP90}
\end{figure}
\FloatBarrier
\begin{figure}[!htbp]
    \centering
    \includegraphics[width=130mm]{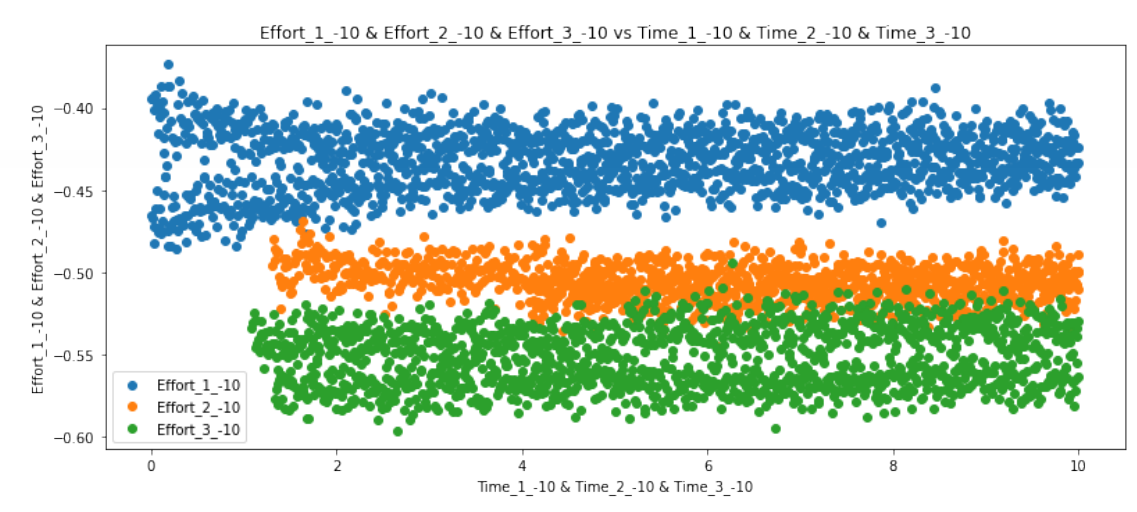}
    \caption{Static phase: Gravitational force coefficient Plots for ${-10 \degree}$}
    \label{fig:E2SP-10}
\end{figure}
\FloatBarrier

\ref{fig:E2SP10}, \ref{fig:E2SP90} and \ref{fig:E2SP-10} graphs are for the static phase of the experiment. \ref{fig:E2SP10}, \ref{fig:E2SP90} and \ref{fig:E2SP-10} show the Effort feedback when the system is held static at a commanded angle. Each graph contains the Effort feedback at the commanded angle, during the positive and negative cycle, across the three experiments.

Title naming convention of the graph is as follows: \newline
${Effort<Experiment number><degree commanded>}$

In \ref{fig:E2SP-10}, there are two sets of blue plots that can be observed, with an effort offset between them. The blue plots represent the effort feedback measured by the system when it is held statically at a position of ${10 \degree}$ during experiment 1. The band on the upper region (having a higher value of effort feedback), is the effort feedback measured when held statically at ${10 \degree}$ during the positive half of the dynamic phase cycle. The band on the lower region  (having a lower value of effort feedback), is the effort feedback measured when held statically at ${10 \degree}$ during the negative half of the dynamic phase cycle. The orange and green bands represent the effort feedback measured by the system when it is held statically at a position of ${10 \degree}$ during experiment 2 and 3 respectively. The time axis for the plots start from 0, in order to make it easier to visualize the readings and draw inference.

In \ref{fig:E2SP90}, similar plots are plotted to represent the effort feedback measured by the system when it is held statically at a position of ${90 \degree}$ across the three experiments.

In \ref{fig:E2SP-10}, similar plots are plotted to represent the effort feedback measured by the system when it is held statically at a position of ${-10 \degree}$ across the three experiments.

There are a few observations from the graphs, which are summarised below
\begin{itemize}
    \item A constant offset in the effort feedback readings of the system is observed when the system is held statically during the positive and negative half cycle across the three experiments.
    \item This offset appears to be constant across the three experimental trials, but varies with the commanded angle at which the system is held static.
    \item This offset may be attributed to the extra effort sensed during the positive half cycle as a result of the commanded motion working against the torque due to gravity.
    \item The constant nature of the effort offset indicates that it is caused due to the hysteresis of the actuator. 
\end{itemize}

The data is collected for the static phase of the experiment for all the angles ranging from ${10 \degree}$ to ${90 \degree}$ with increments of ${10 \degree}$ across both the positive and negative cycles for the three experimental trials. A plot of the static effort feedback and sine of the angle at which the pendulum system is held is plotted below.

\begin{figure}[!htbp]
    \centering
    \includegraphics[width=150mm]{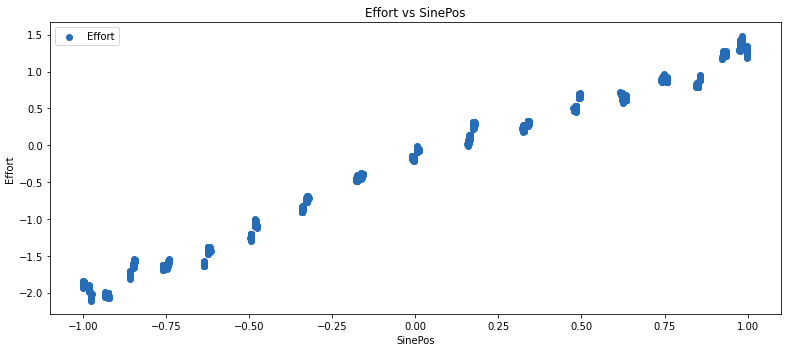}
    \caption{Effort Feedback vs Sine of Position feedback}
    \label{fig:EFvsSPF}
\end{figure}
\FloatBarrier

In \ref{fig:EFvsSPF}, a linear relationship can be observed between the effort feedback and the sine of the angle, as is expected from the dynamical equation.

\begin{figure}[!htbp]
    \centering
    \includegraphics[width=90mm]{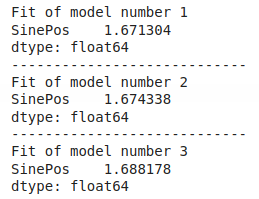}
    \caption{Gravitational coefficient experiment regressed coefficients}
    \label{fig:GCCoeff}
\end{figure}

Effort feedback for all the recorded angles is used as the dependant variable and sine of the angle at which the system is held static is used as the independent variable to set up a simple Least Squares Regression model for each experiment trial. The result of the regression model for each experiment trial is shown in \ref{fig:GCCoeff}. The SinePos coefficient corresponds to the quantity ${G_{c}}$ as given in \ref{eqn:PSDEqnConstants}.

\begin{figure}[!htbp]
    \centering
    \includegraphics[width=150mm]{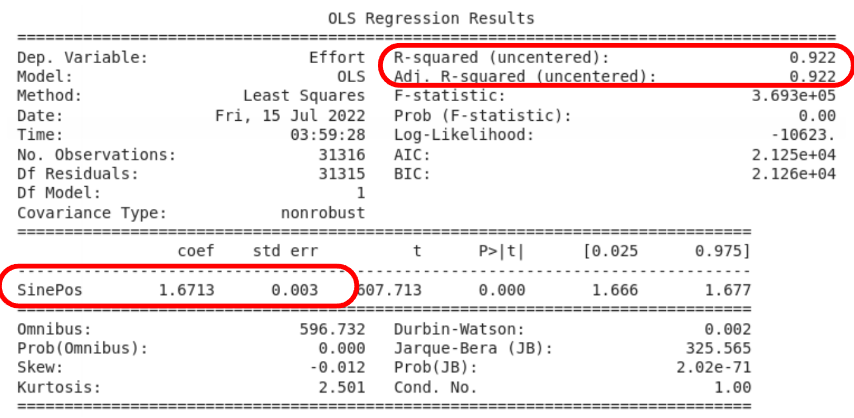}
    \caption{Gravitational coefficient OLS}
    \label{fig:GCOLS}
\end{figure}
\FloatBarrier
\FloatBarrier

An average of the coefficient across the 3 experiments is computed and the gravitational force coefficient is obtained as below

${\mathbf{G_{c}} : 1.678 Nm}$

\section{Simulation setup for dynamical system}

\subsection{Open AI Gym Environment Setup}
Open AI Gym \cite{brockman_2016_openai} is an open source python library that provides various open source environments for testing reinforcement learning algorithms. These environments consist of a standard set of functions and variables which have been described in \cite{brockman_2016_openai}. The subsequent discussion builds on top of these conventions. Some of the environments include cart-pole, acrobot, lunar lander etc. 

\begin{figure}[!htbp]
    \centering
    \includegraphics[width=150mm]{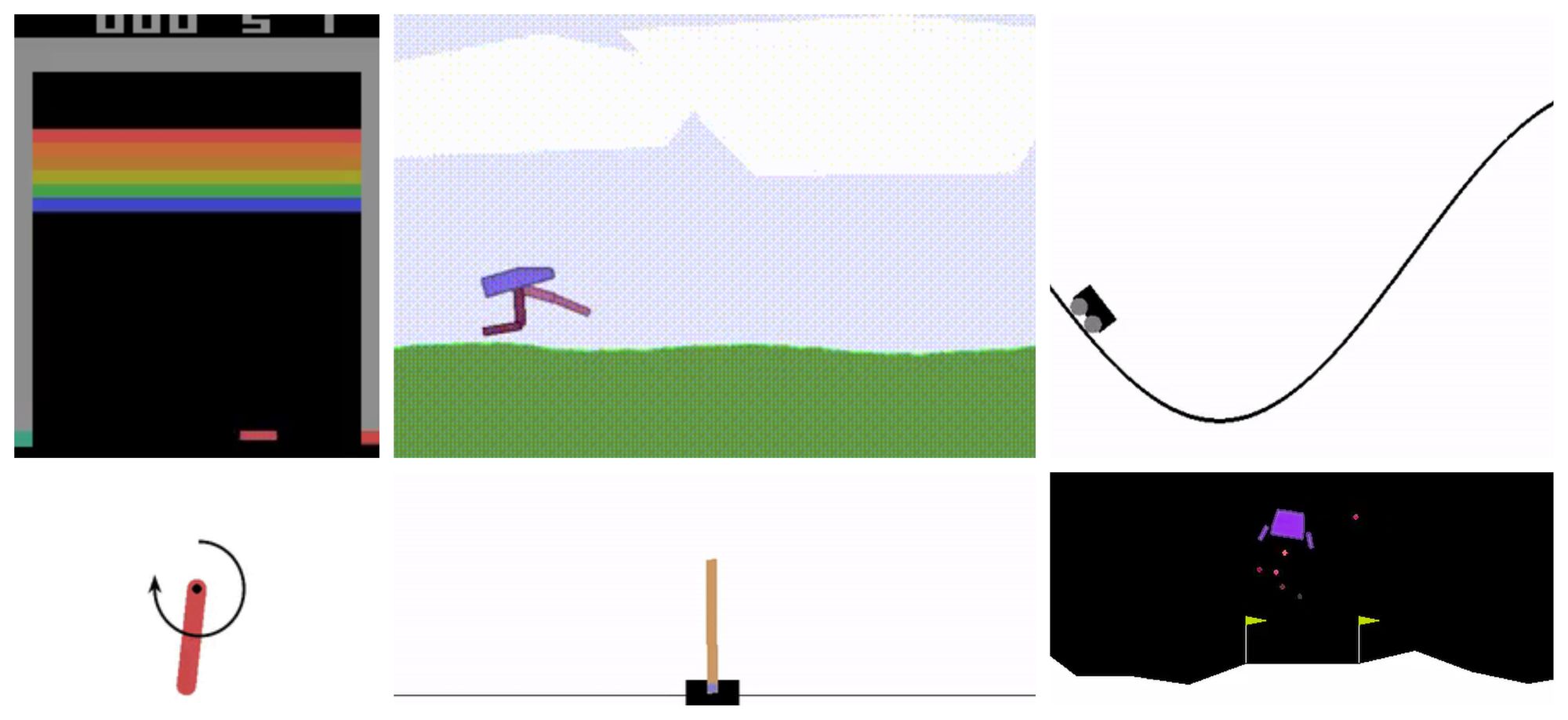}
    \caption{OpenAI Gym environments}
    \label{fig:GymEnv}
\end{figure}
\FloatBarrier

However, these environments suffer from a few disadvantages that render them inconvenient for testing optimal control algorithms, without requiring much change when shifting from reinforcement learning algorithms to optimal control algorithms. These disadvantages are described below
\begin{itemize}
    \item The environments do not make use of standard control theory convention for representing state vectors.
    \item The environments use Euler integration with a step size of ${0.02sec}$, which turned out to be coarse in order to simulate optimal control algorithms effectively.
    \item The environments are highly oriented towards testing reinforcement learning algorithms. 
    \item The action (control input) on most of the control problem environments are of type discrete and do not model continuous torque required to model optimal control algorithms. 
\end{itemize}

In order to address the above concerns, changes were made to the appropriate environment functions to enable easier testing of algorithms from optimal control theory. The specifics of these changes are described below

\begin{itemize}
    \item A step size of ${0.002sec}$ is used to integrate the dynamics of the system. Euler integration is used as the integrator. The above settings enabled more accurate simulation of the dynamics of the system.
    \item The state vector, which includes the observation returned by the environment, is changed to follow standard control convention.
    \item The action space of the environment is changed to Box type (Open AI Gym class to represent continuous vector spaces), to better represent the continuous nature of control inputs that will be generated by control algorithms.
    \item The rendering of the environment is changed to incorporate a custom input from the user detailing the screen size to display the simulation.
    \item A rotation matrix is used to calculate the display location of the subsequent state of the dynamical system.
    \item A function to plot the evolution of the state variables is added.
\end{itemize}

\subsection{Pendulum environment setup}
The Open AI Gym library provides an inverted pendulum setup. However, the open source implementation is not used for the below reason
\begin{itemize}
    \item The open source implementation focuses on an RL algorithm to swing up the pendulum to its unstable point and balance it there. The focus of this report is rather on designing and validating the performance of optimal control algorithms in simulation by becnhmarking it with the hardware response.
\end{itemize}

\begin{figure}[!htbp]
    \centering
    \includegraphics[width=80mm]{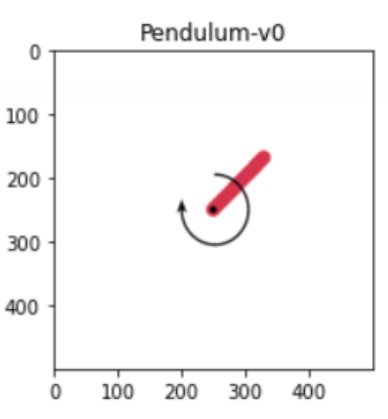}
    \caption{OpenAI Gym Pendulum-v0}
    \label{fig:PendulumGymEnv}
\end{figure}
\FloatBarrier

A new environment is setup taking into consideration the above changes. The standard environment function changes have been described above. The specific details of the dynamics of the system have been detailed below.

\begin{itemize}
    \item The dynamical equation of the pendulum system \ref{eqn:PSDEqnConstants} is set up with the Inertial, damping and Gravitational force coefficient obtained via experiments that have been described in Chapter 4.
    \item Equation \ref{eqn:PSDEqnConstants} is used to calculate the acceleration component, which is then used to update the state of the system using Euler integration with a step size of ${0.002sec}$
\end{itemize}

\begin{figure}[!htbp]
    \centering
    \includegraphics[width=120mm]{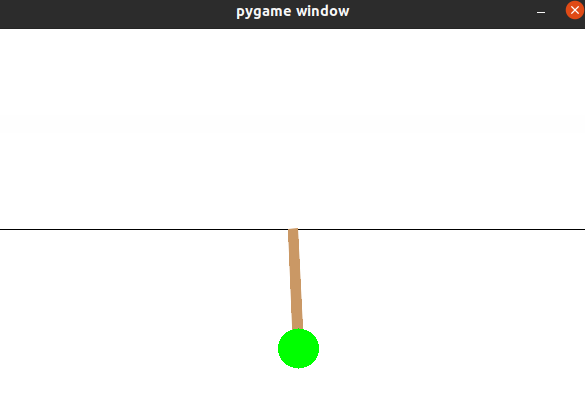}
    \caption{Pendulum Pygame simulation setup}
    \label{fig:PendulumPGSS}
\end{figure}
\FloatBarrier

\section{Optimal control: LQR}

This chapter details the use of the optimal control technique: Linear Quadratic Regulator in the case of the pendulum system. The control objective is to regulate the system at its unstable fixed point of ${\mathbf{x} = \begin{bmatrix}
    \pi & 0
    \end{bmatrix}}$. 
A LQR controller is designed and tested in the presence of white noise in simulation. The same LQR controller is then tested on hardware in the presence of white noise and the system response is recorded.

A stable dynamical system is characterised by a state transition matrix whose eigenvalues are all negative. The eigenvalues are equivalent to the poles of the dynamical system. Hence, an eigenvalue with a large negative magnitude represents a fast decaying system. 

The subsequent discussion makes use of this concept in order to select the cost matrices ${\mathbf{Q}}$ and ${\mathbf{R}}$ which lead to a desired controller gain ${K_{r}}$ as described in Section 2.2.1.

The state transition equation of the dynamical system \ref{eqn:PSDEqn} can be obtained by linearizing the dynamics of the simple pendulum system about the unstable fixed point of ${\mathbf{x} = \begin{bmatrix}
    \pi & 0
    \end{bmatrix}}$.
    
The linearized dynamics about the unstable fixed point can be obtained by evaluating the Jacobian of the vector ${\dot{\mathbf{x}}}$ with respect to the vector ${\mathbf{x}}$, where ${\mathbf{x}}$ represents the state vector of the pendulum system and ${\dot{\mathbf{x}}}$ represents its derivative.

From \ref{eqn:PSDEqn}, we can obtain 
\begin{center}
   ${\ddot{\theta}}$ =${\frac{u - b\dot{\theta} -mgl_c\sin\theta}{ml_c^2 + I_c}}$ 
\end{center}

The Jacobian matrices ${A_{lin}}$ and ${B_{lin}}$ are as below

\begin{center}
    ${A_{lin}}$=${\begin{bmatrix}
                \frac{\partial \dot{\theta}}{\partial \theta} & \frac{\partial \dot{\theta}}{\partial \dot{\theta}}\\
                \frac{\partial \ddot{\theta}}{\partial \theta} & \frac{\partial \ddot{\theta}}{\partial \dot{\theta}}
                \end{bmatrix}}$
\end{center}

\begin{center}
    ${B_{lin}}$=${\begin{bmatrix}
\frac{\partial \dot{\theta}}{\partial u} \\ \frac{\partial \ddot{\theta}}{\partial u}
\end{bmatrix}}$
\end{center}

Differentiating ${\ddot{\theta}}$ with respect to ${\theta}$, ${\dot{\theta}}$ and ${u}$ gives 

\begin{center}
    ${\frac{\partial \ddot{\theta}}{\partial \theta}}$ = ${\frac{-mgl_c\cos \theta}{ml_c^2 + I_c}}$ 
\end{center}

\begin{center}
    ${\frac{\partial \ddot{\theta}}{\partial \dot{\theta}}}$ = ${\frac{-b}{ml_c^2 + I_c}}$
\end{center}

\begin{center}
    ${\frac{\partial \ddot{\theta}}{\partial u}}$ = ${\frac{1}{ml_c^2 + I_c}}$ 
\end{center}

The obtained expressions are evaluated at the unstable point ${\mathbf{x} = \begin{bmatrix}
    \pi & 0
    \end{bmatrix}}$, by substituting the values of ${\theta}$ and ${\dot{\theta}}$ correspondingly. This results in the below expression

\begin{center}
    ${\frac{\partial \ddot{\theta}}{\partial \theta}|_{\theta=\pi, \dot{\theta}=0}}$ = ${\frac{mgl_c}{ml_c^2 + I_c}}$
\end{center}    

\begin{center}
    ${\frac{\partial \ddot{\theta}}{\partial \dot{\theta}}|_{\theta=\pi, \dot{\theta}=0}}$ = ${\frac{-b}{ml_c^2 + I_c}}$
\end{center}

\begin{center}
    ${\frac{\partial \ddot{\theta}}{\partial u}|_{\theta=\pi, \dot{\theta}=0}}$ = ${\frac{1}{ml_c^2 + I_c}}$ 
\end{center}

The above expression can be further abstracted to the form 
\begin{center}
  ${\frac{\partial \ddot{\theta}}{\partial \theta}|_{\theta=\pi, \dot{\theta}=0}}$ = ${\frac{G_c}{M_c}}$  
\end{center}

\begin{center}
    ${\frac{\partial \ddot{\theta}}{\partial \dot{\theta}}|_{\theta=\pi, \dot{\theta}=0}}$ = ${\frac{-b}{M_c}}$
\end{center}

\begin{center}
    ${\frac{\partial \ddot{\theta}}{\partial u}|_{\theta=\pi, \dot{\theta}=0}}$ = ${\frac{1}{M_c}}$ 
\end{center}

The values of ${M_c}$, ${b}$ and ${G_c}$ have been obtained in Section: Experimental determination of parameters of system dynamics. Substituting the values, the expressions evaluate to
\begin{center}
  ${\frac{\partial \ddot{\theta}}{\partial \theta}|_{\theta=\pi, \dot{\theta}=0}}$ = ${30.51}$
\end{center}

\begin{center}
    ${\frac{\partial \ddot{\theta}}{\partial \dot{\theta}}|_{\theta=\pi, \dot{\theta}=0}}$ = ${-214}$
\end{center}

\begin{center}
    ${\frac{\partial \ddot{\theta}}{\partial u}|_{\theta=\pi, \dot{\theta}=0}}$ = ${18.18}$ 
\end{center}

The Jacobian matrices evaluated at the unstable fixed point are as below
\begin{center}
    ${A_{lin}}$ =${\begin{bmatrix}
                0 & 1\\
                30.51 & -214
                \end{bmatrix}}$
\end{center}

\begin{center}
    ${B_{lin}}$=${\begin{bmatrix}
0 \\ 18.18
\end{bmatrix}}$
\end{center}

Hence, the linearized dynamics of the system about the unstable fixed point is obtained as below

\begin{center}
    ${\dot{\mathbf{x}}}$ = ${\begin{bmatrix}
                0 & 1\\
                30.51 & -214
                \end{bmatrix}}$${\mathbf{x}}$ + ${\begin{bmatrix}
0 \\ 18.18
\end{bmatrix}}$${u}$
\end{center}

This can be expanded as below
\begin{center}
    ${\begin{bmatrix}
        \dot{\theta} \\
        \ddot{\theta}
    \end{bmatrix}}$ = ${\begin{bmatrix}
                0 & 1\\
                30.51 & -214
                \end{bmatrix}}$ ${\begin{bmatrix}
        \theta - \pi \\
        \dot{\theta} - 0
    \end{bmatrix}}$ + ${\begin{bmatrix}
0 \\ 18.18
\end{bmatrix}}$ ${\begin{bmatrix}
        0 \\
        u - 0
    \end{bmatrix}}$
\end{center}

The poles of the system are equivalent to the eigenvalues of the matrix ${A_{lin}}$ in an open loop system with no feedback. 

\subsection{Eigenvalues with LQR feedback}
When the LQR feedback with a controller gain of ${K_{r}}$ is fed to the system, the state transition equation can be updated as below

\begin{equation} \dot{\mathbf{x}} = (A_{\text{lin}} - K_{r}B_{\text{lin}})\mathbf{x}\end{equation}

The eigenvalues of the matrix ${(A_{\text{lin}} - K_{r}B_{\text{lin}})}$ are equivalent to the poles of the system and hence describes the stability of the system. The eigenvalues for the above matrix is calculated for different combinations of cost matrices ${\mathbf{Q}}$ and ${\mathbf{R}}$, and the results are in the table below. The columns in the table refer to the corresponding values in the cost matrices

\begin{center}
    ${\mathbf{Q} = \begin{bmatrix}
q11 & q12\\
q21 & q22
\end{bmatrix}}$
\end{center}

\begin{center}
    ${\mathbf{R} = \begin{bmatrix}
r11
\end{bmatrix}}$
\end{center}

\begin{table}[h!]
\centering
\begin{tabular}{|c|c|c|c|c|c|c|c|c|} 
 \hline
 Combination & Feedback & q11 & q12 & q21 & q22 & r11 & Eigenvalue 1 & Eigenvalue 2\\ [0.5ex] 
 \hline \hline
 0 & No Feedback & - & - & - & - & - & 0.14 & -214.14\\ 
 \hline
 1 & LQR Feedback & 1 & 0 & 0 & 0.01 & 0.1 & -0.30 & -214.22 \\
 \hline
 2 & LQR Feedback & 1 & 0 & 0 & 0.1 & 0.1 & -0.30 & -214.91 \\
 \hline
 3 & LQR Feedback & 100 & 0 & 0 & 0.01 & 0.1 & -2.69 & -214.20 \\
 \hline
 4 & LQR Feedback & 100 & 0 & 0 & 0.1 & 0.1 & -2.68 & -214.90 \\ [1ex] 
 \hline
\end{tabular}
\caption{Eigenvalues for LQR response of different cost matrices}
\label{table:Cost Matrix Combinations}
\end{table}

From the table \ref{table:Cost Matrix Combinations}, it can be seen that Combination 0, which is the open loop response of the dynamical system with no LQR feedback, has Eigenvalue 1 with a positive magnitude. The system has one pole in the positive half of the s-plane. This represents an unstable system with a diverging state response.

On the contrary, the remaining combinations, which are the closed loop system response with the LQR controller, have all negative Eigenvalues. All the poles of the system are in the negative half of the s-plane. This represents a stable dynamical system.

\subsection{Time constant of the system}
From the table \ref{table:Cost Matrix Combinations} of eigenvalues for different cost matrices, it can be seen that Combination 3 has Eigenvalue 1 with a larger negative magnitude when compared to the other combinations. This implies a faster stabilising system. For this reason, Combination 3 is chosen for further analysis using the simulation and hardware. 

The optimal control gain with the cost matrices from Combination 3 is given below \newline
${K_{r} = \begin{bmatrix}
        33.345 & 0.159
    \end{bmatrix}}$

For a dynamical system which decays exponentially, the time constant of the system can be represented by the below equation
\begin{equation}
\label{eqn:TimeConstantExpDecay}
\tau = \frac{1}{|v|}
\end{equation}

where ${|v|}$ represents the absolute value of the eigenvalue of the system. For Combination 3, the time constant of the dynamical system is calculated below \newline
${\tau_{Combination 3} = \frac{1}{2.69} = 0.3717 s^{-1}}$

\subsection{Simulation: LQR response limit check}
\begin{figure}[!htbp]
    \centering
    \includegraphics[width=150mm]{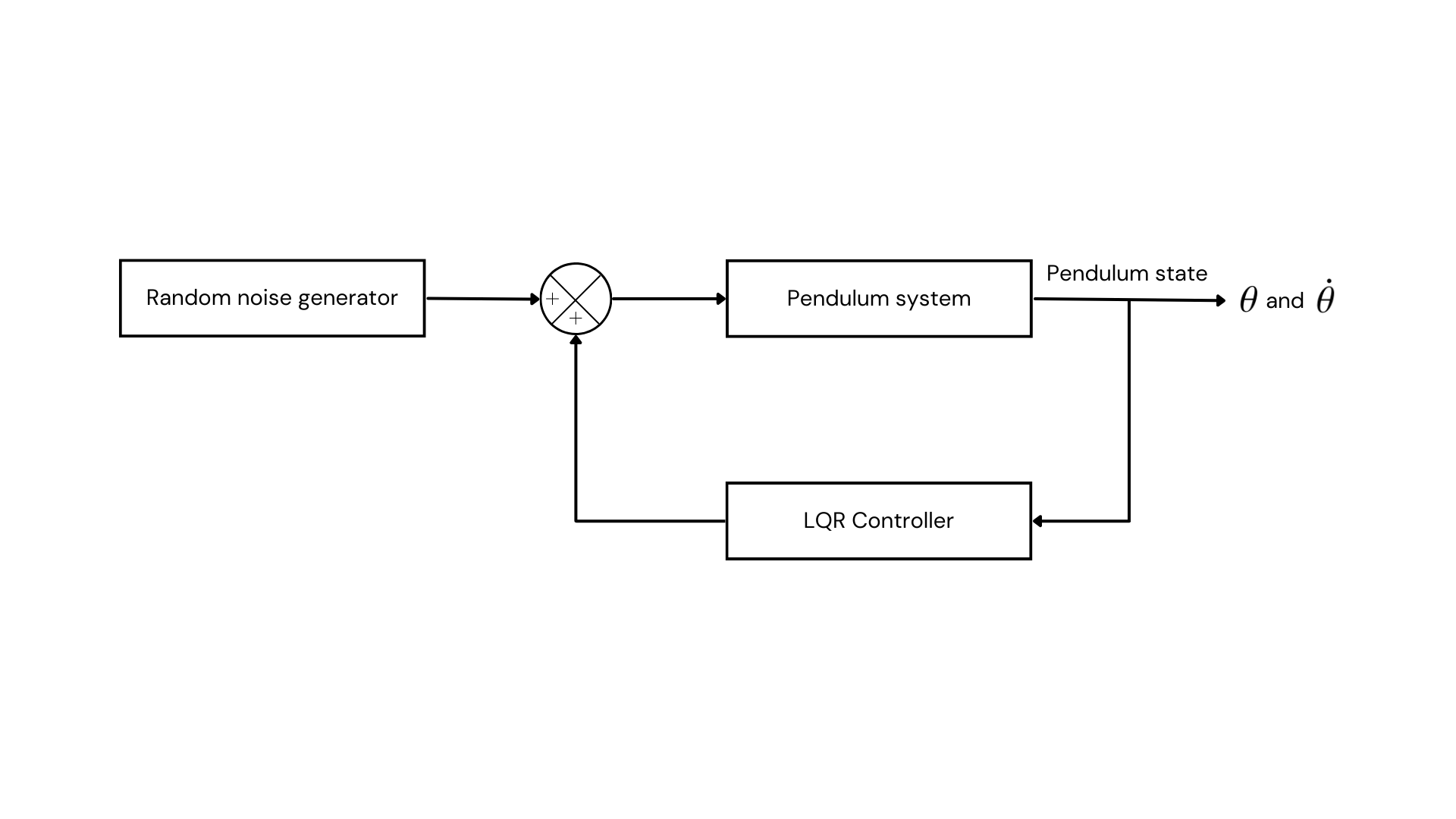}
    \caption{Block diagram of pendulum system}
    \label{fig:BD}
\end{figure}

In order to examine if the system response is within the safe limits to be tested on hardware, the LQR response is tested in simulation by applying a disturbance in the form of white noise. The noise is generated using the python function ${random.uniform}$ present in the ${random}$ library. A random number ${N_{r}}$ is generated in the range of ${N_{r} \in [-2.5, 2.5] Nm}$. ${N_{r}}$ is the torque applied to the system in the form of noise. The LQR controller is kept active throughout the simulation. The system state output after applying the noise is fed to the LQR controller. The LQR controller then generates a control input to regulate the system at its unstable fixed point. This forms a closed loop system, the block diagram of the same is shown in \ref{fig:BD}.

\FloatBarrier

The system is simulated for a duration of ${5s}$ and the data for one simulation is recorded and plotted below

\begin{figure}[!htbp]
    \centering
    \includegraphics[width=150mm]{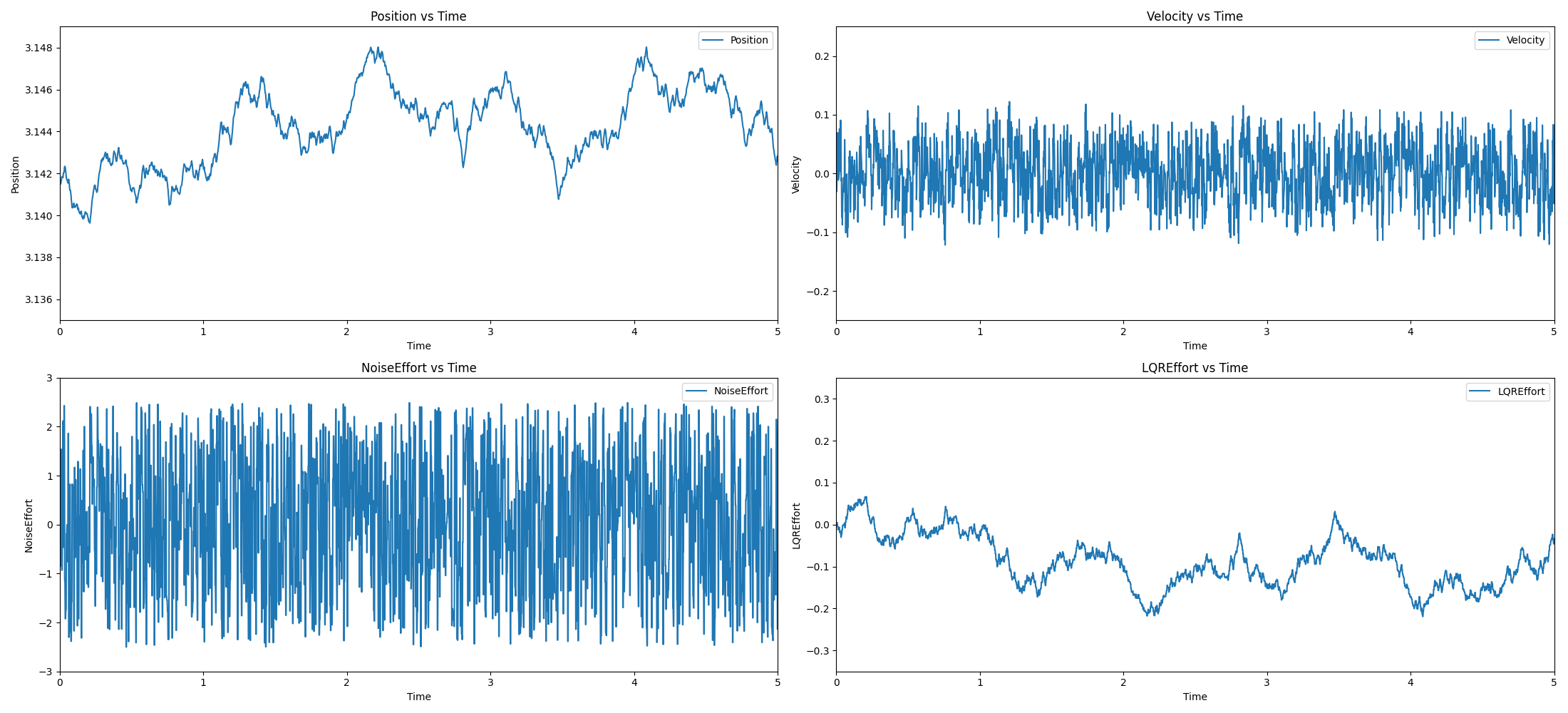}
    \caption{Simulation 1: Pendulum system LQR response to noise}
    \label{fig:LQRSIM1}
\end{figure}
\FloatBarrier

From the plot \ref{fig:LQRSIM1}, the response of the LQR controller can be noted. The LQR controller is able to maintain the position and velocity of the system close to the value of the unstable fixed point. The maximum absolute amplitude of torque exerted by the LQR controller is around ${|\tau_{LQRmax}| \approx 0.25Nm}$. This value is well within the permissible inputs to the HEBI motor. The LQR controller is now tested on hardware.

\subsection{Hardware: LQR response to noise}
\begin{figure}[!htbp]
    \centering
    \includegraphics[width=150mm]{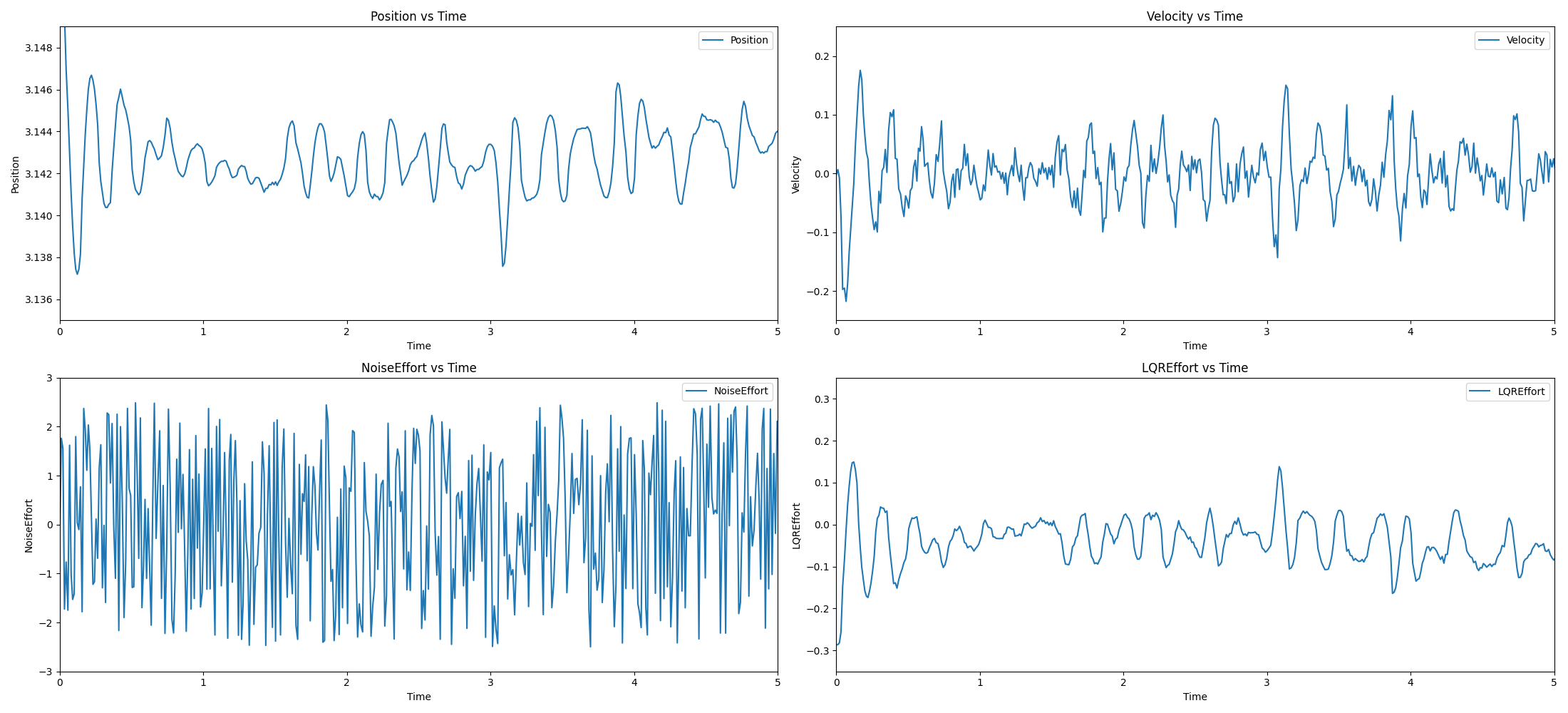}
    \caption{Experiment 1: Pendulum system LQR response to noise}
    \label{fig:LQREXP1}
\end{figure}

In order to test the LQR controller response on hardware, the pendulum system was brought to the state ${\mathbf{x} = \begin{bmatrix}
    \pi & 0
    \end{bmatrix}}$ using Strategy 3 by commanding a smooth position and velocity curve as discussed in the Section: Determination of Gravitational coefficient of the system in Chapter: Experimental determination of parameters of system dynamics. The LQR controller is then activated after reaching the state. Disturbance in the form of white noise as described in the previous section is applied to the hardware system in the form of torque commanded to HEBI and the response of the LQR controller and the system is recorded. The control system is a feedback loop similar to \ref{fig:BD}. The plots from one experiment is shown in \ref{fig:LQREXP1}

\FloatBarrier

The system response from hardware and simulation are plotted together as below

\begin{figure}[!htbp]
    \centering
    \includegraphics[width=150mm]{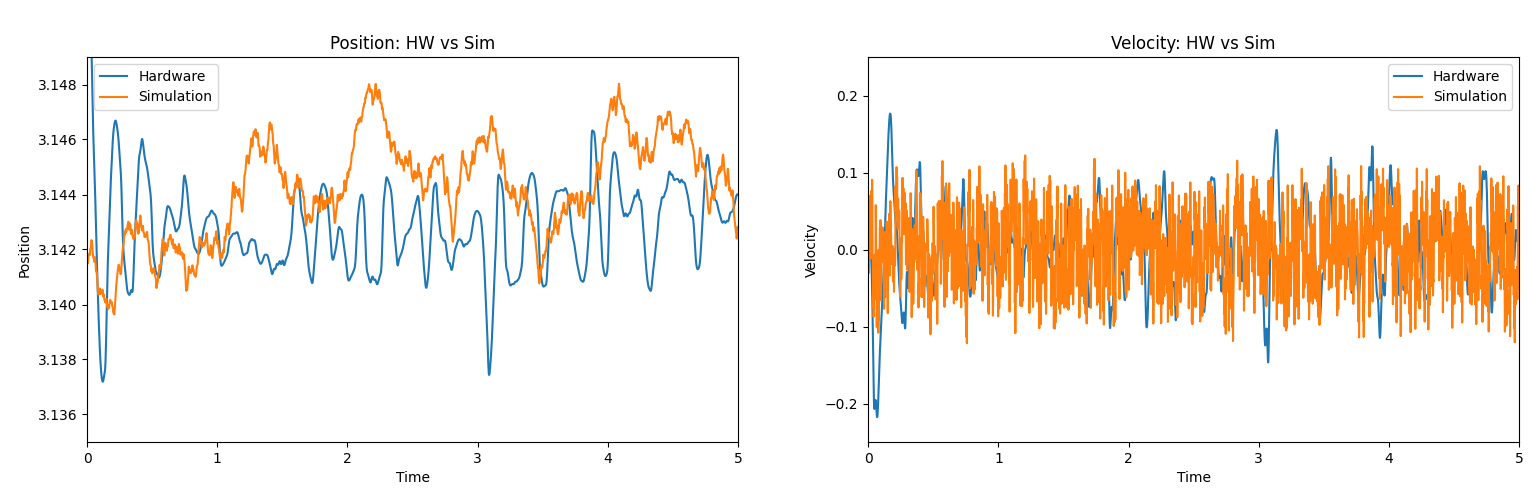}
    \caption{HW vs SIM: Pendulum system LQR response to noise}
    \label{fig:LQRSIMEXP1}
\end{figure}
\FloatBarrier

From the plot \ref{fig:LQRSIMEXP1}, the response of the system matches closely with the response obtained from simulation. 

The non scaled version of the plot \ref{fig:LQRSIMEXP1} is plotted below

\begin{figure}[!htbp]
    \centering
    \includegraphics[width=150mm]{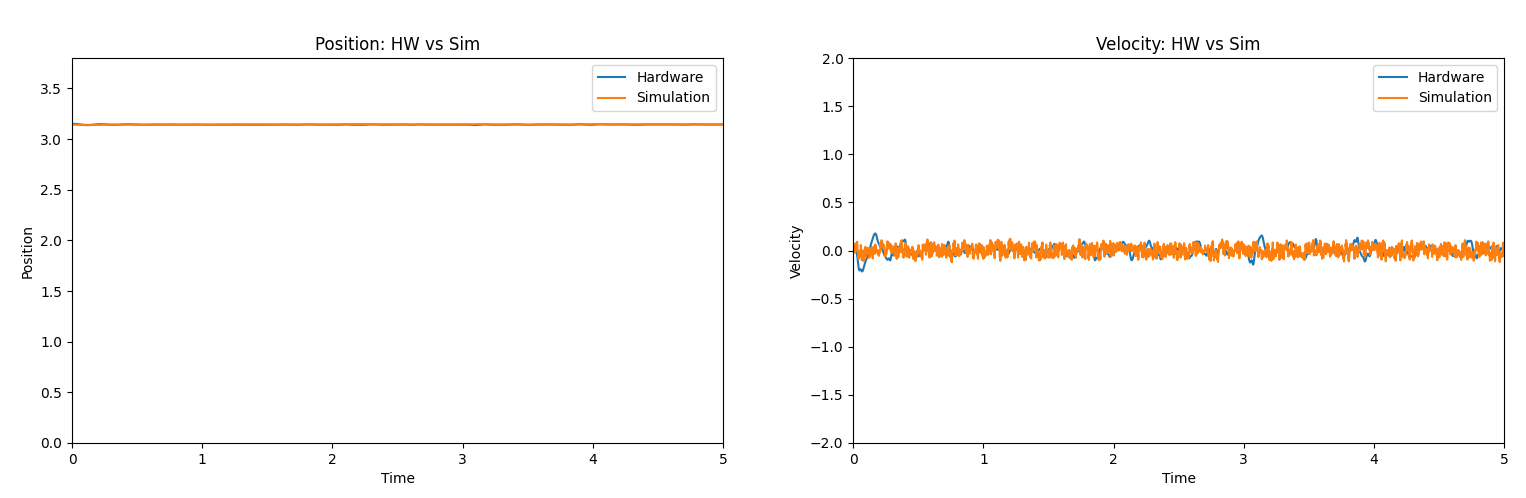}
    \caption{HW vs SIM: Pendulum system LQR response to noise unscaled}
    \label{fig:LQRSIMEXPSCALED1}
\end{figure}
\FloatBarrier

From the plot \ref{fig:LQRSIMEXPSCALED1}, it can be observed that the designed LQR controller is able to regulate the target system at the unstable fixed point even under the presence of a high magnitude noise. It is also observed that the system response from simulation matches the one from hardware closely.

A comparison of the system response between hardware and simulation is done in the next chapter. 

\section{Analysis of system response: Hardware and Simulation}

\subsection{Preparation of system response data}
From the previous chapter, the system response with the LQR controller in the presence of white noise has been recorded from both sources, simulation and hardware. The data obtained from simulation has a time step of ${0.002sec}$ which is the Euler integration time step, or a frequency of ${500Hz}$. However, the data obtained from hardware has a frequency of ${100Hz}$. 

In order to do an analysis on the simulation system response data and hardware system response data, the time steps of both the data need to match. This is done by interpolating the system response obtained from hardware. For the purposes of interpolation, the Python ${scipy}$ library is used. A cubic interpolation of the hardware system response at the same time stamps of the simulation data is done and a corresponding interpolated hardware system response is obtained.

\begin{figure}[!htbp]
    \centering
    \includegraphics[height=35mm, width=150mm]{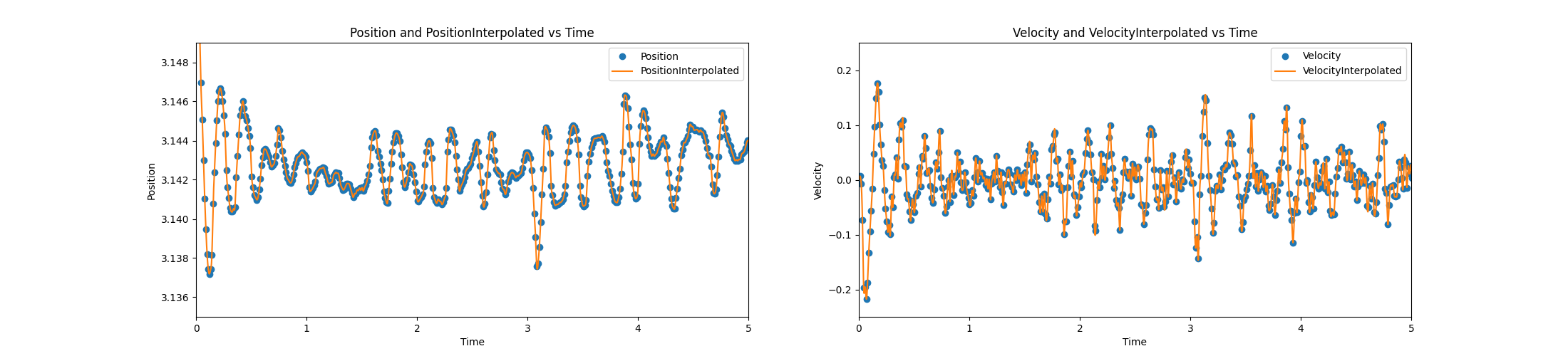}
    \caption{Experiment 1: Interpolated pendulum system LQR response to noise}
    \label{fig:LQREXPSIMInter1}
\end{figure}

The plot \ref{fig:LQREXPSIMInter1} compares the original and the interpolated hardware response from the experiment mentioned in Chapter 6: Optimal Control: LQR under Section 6.4 Hardware: LQR response to noise. 

\FloatBarrier

\subsection{Analysis of system response data}
\begin{figure}[!htbp]
    \centering
    \includegraphics[width=150mm]{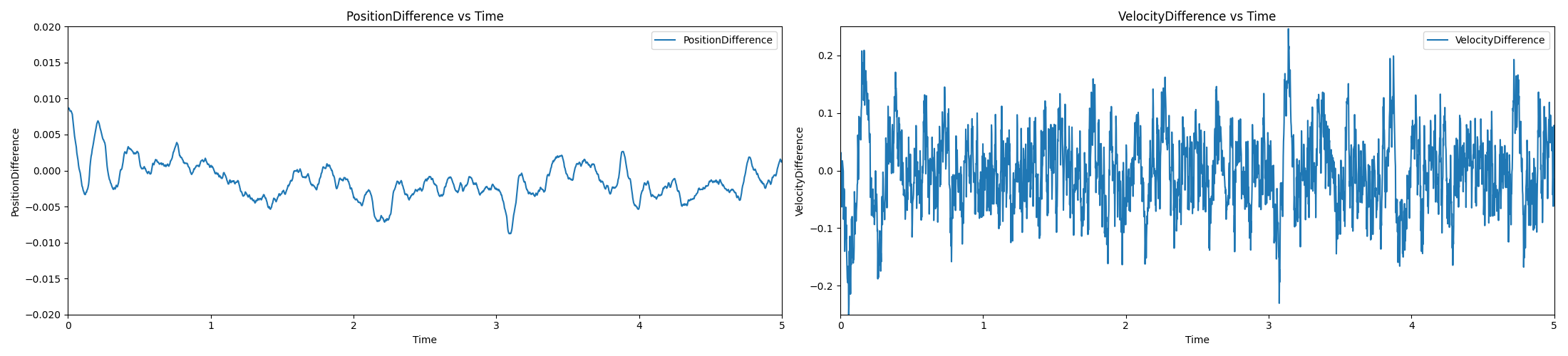}
    \caption{Difference between simulation and hardware state response}
    \label{fig:LQRDiff1}
\end{figure}

The external disturbance to the experiment and simulation is in the form of white noise. To examine if the simulation is able to model the hardware response under the LQR controller, a difference of the state response obtained from simulation and the interpolated hardware response is computed. A statistical inference on the computed difference is done. \ref{fig:LQRDiff1} shows the plot of the difference in system response.

\FloatBarrier

A statistical inference on the above plots yields the below result
\begin{figure}[!htbp]
    \centering
    \includegraphics[width=90mm]{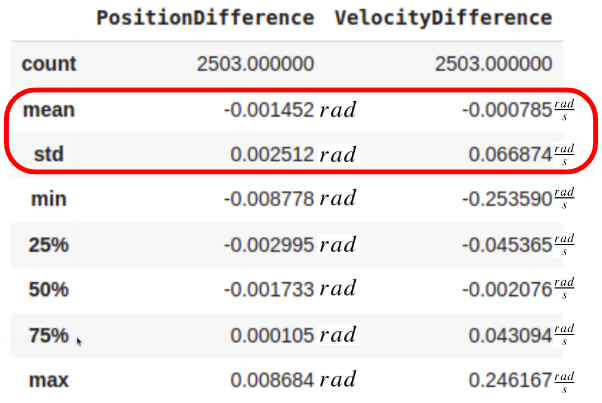}
    \caption{Statistical inference on difference between simulation and hardware state response}
    \label{fig:LQRDiffInference1}
\end{figure}
\FloatBarrier

From \ref{fig:LQRDiffInference1}, it can be seen that the mean of both the position and velocity difference between the simulated and interpolated hardware response is very close to ${0 rad}$ and ${0 rad/s}$ respectively. This states that the hardware response and simulation response do not have any offset difference between their responses. The standard deviation of both the position and velocity difference between the simulated and interpolated hardware response is also very close to ${0 rad}$ and ${0 rad/s}$ respectively. This states that the hardware response and simulation response data points are similar to each other. Combining these two results, it can be validated that the simulation of an approximated system model is able to model the hardware response in a closed loop system with a LQR controller and a disturbance in the form of white noise.

\section{Conclusion}
This report has validated that simulation of approximated system models can be used to model the hardware system response under a controller. From the comparative statistical analysis of the system response in simulation and on hardware with a LQR controller subjected to white noise, it has been shown that the simulation is able to model the dynamics of the controller's response.

As robots and control algorithms evolve in complexity, simulation is a vital tool to model a robot's behavior and test different control algorithms before deploying them in real world environment. This can result in huge savings in finances, time and also help test the edge cases which may be dangerous to test in real world environments. 

This report looked at a pendulum system, but the experiments designed in the report can be used to develop approximated system models of more complex dynamical systems. These approximated models can be used to design and test control algorithms in simulation.

\subsection{Future work scope}
The conditions, components and control algorithms used in the report can be changed to many different possibilities. This opens the possibility to future work such as

\begin{itemize}
    \item Using Reinforcement learning algorithms to simulate model free dynamical systems and compare with hardware response
    \item Using underactuated systems such as a cartpole, acrobot etc as the dynamical system of choice
    \item Imparting the disturbance physically on the hardware system
    \item Comparing the system response between hardware and simulation starting from a random point to the regulated point
\end{itemize}

\printbibliography

\end{document}